\documentclass[10pt,twocolumn,letterpaper]{article}

\usepackage[pagenumbers]{cvpr} 

\usepackage{url}
\usepackage{amsmath}

\usepackage{booktabs}
\usepackage{xcolor, colortbl}
\usepackage{multirow} 
\usepackage{graphicx} 
\usepackage{caption}  
\usepackage{subcaption}  
\usepackage{xcolor}  
\usepackage{pifont}  
\definecolor{almond}{rgb}{0.94, 0.87, 0.8}
\usepackage{graphics}
\usepackage{wrapfig}
\usepackage{tabularx}
\usepackage{float}
\usepackage[export]{adjustbox}  
\usepackage[numbers, sort&compress]{natbib}

\definecolor{cvprblue}{rgb}{0.21,0.49,0.74}
\usepackage[pagebackref,breaklinks,colorlinks,citecolor=cvprblue]{hyperref}

\title{Grounding Everything: Emerging Localization Properties \\ in Vision-Language Transformers}

\author{%
    Walid Bousselham$^{1,2}$ \quad
    Felix Petersen$^3$  \quad
    Vittorio Ferrari$^{4}$  \quad
    Hilde Kuehne$^{1,2,5}$ \\
    \kern-1.1em\small{
    $^1$University of Bonn,
    $^2$Goethe University Frankfurt,
    $^3$Stanford University, 
    $^4$Synthesia.io, 
    $^5$MIT-IBM Watson AI Lab
    } \\
}

\begin{document}
\maketitle
\begin{abstract}
Vision-language foundation models have shown remarkable performance in various zero-shot settings such as image retrieval,  classification, or captioning. But so far, those models seem to fall behind when it comes to zero-shot localization of referential expressions and objects in images. 
In this paper, we show that pretrained vision-language (VL) models allow for zero-shot open-vocabulary object localization without any fine-tuning. To leverage those capabilities, we propose a Grounding Everything Module (GEM) that generalizes the idea of value-value attention introduced by CLIPSurgery~\citep{li2023clipsurgery} to a self-self attention path. We show that the concept of self-self attention corresponds to clustering, thus enforcing groups of tokens arising from the same object to be similar while preserving the alignment with the language space. To further guide the group formation, we propose a set of regularizations that allows the model to better generalize across datasets and backbones. 
We evaluate the proposed GEM framework on various benchmark tasks and datasets for semantic segmentation. It shows that GEM not only outperforms other training-free open-vocabulary localization methods, but also achieves state-of-the-art results on the recently proposed OpenImagesV7 large-scale segmentation benchmark. \footnote{Code available at \url{https://github.com/WalBouss/GEM}}\footnote{Demo available at \url{https://huggingface.co/spaces/WalidBouss/GEM}}
\end{abstract}    
\section{Introduction}
\label{sec:intro}
\begin{figure}[ht]
\centering \label{fig:teaser-fig}
     \includegraphics[width=0.47\textwidth]{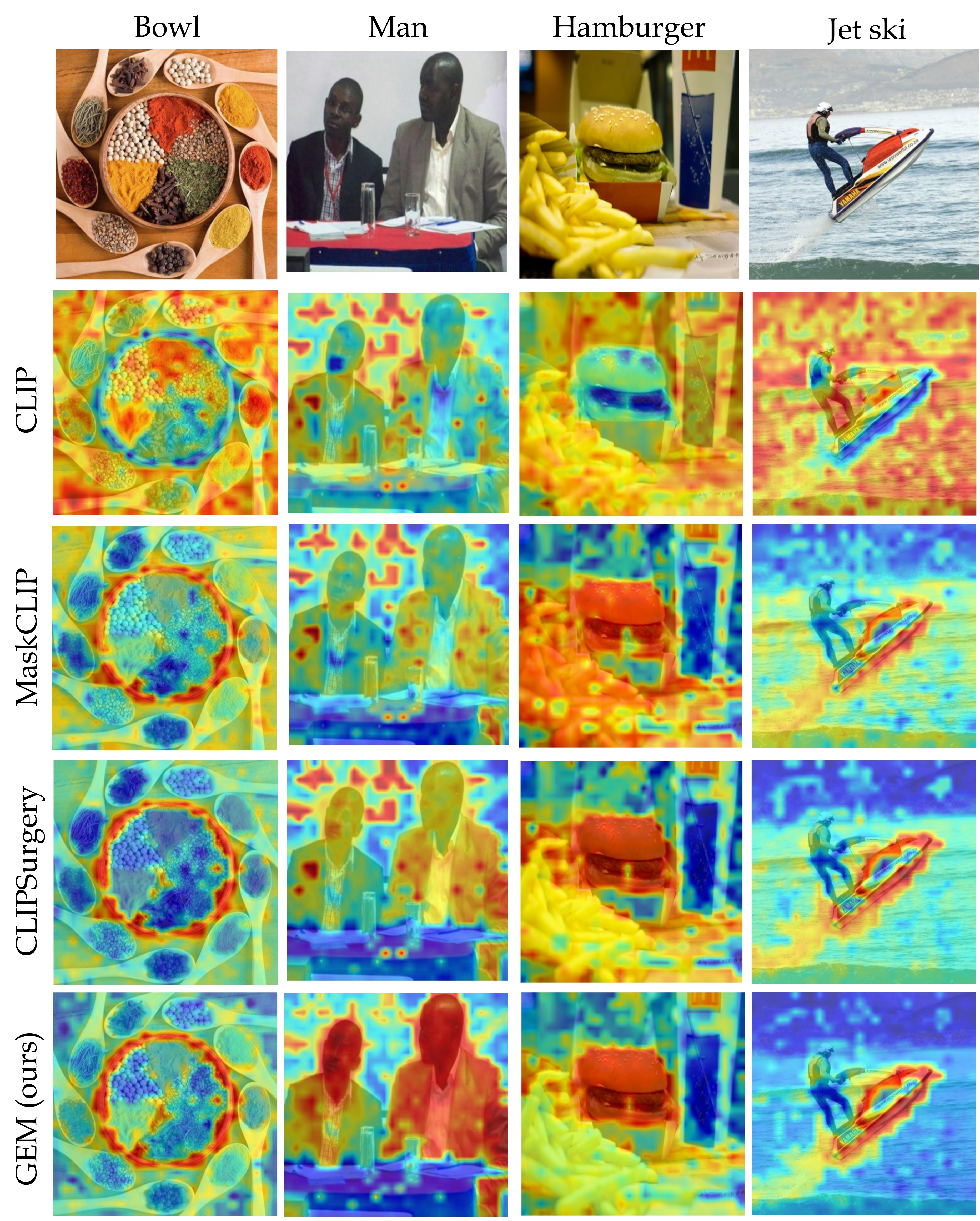}
     \caption{Qualitative results of training-free methods: given a text prompt, the similarity of each image token with the prompt is calculated (red:high, blue:low). The proposed GEM method provides improved grouping and alignment compared to other approaches.}
     \label{fig:teaser}
\end{figure}
Vision-language models, trained on large-scale web-based datasets such as WIT-400M~\citep{radford2021learning}, LAION400M~\citep{schuhmann2022laion}, or metaclip-400M~\citep{xu2023demystifying} with image-text supervision only, have so far shown a remarkable set of capabilities. These models such as CLIP~\citep{radford2021learning}, OpenCLIP~\citep{schuhmann2022laion}, BLIP~\citep{li2022blip}, or recently MetaCLIP~\citep{xu2023demystifying} exhibit the ability to generalize to a broad range of downstream tasks like zero-shot image classification~\citep{radford2021learning, jia2021scaling, cherti2023reproducible}, visual question answering~\citep{khan2022weakly}, action recognition~\citep{yuan2021florence, yu2022coca}, image captioning ~\citep{li2022blip, li2019visualbert},  and view synthesis~\citep{jain2021putting}. 
However, models trained with image-level objectives such as contrastive loss, image-text matching, or image captioning struggle to maintain their zero-shot capabilities for tasks related to visual localization. Even worse, when prompting such models for e.g., specific objects, they show an inverse vision-language relation, thus, image patches showing the object have usually a larger distance from the prompt embedding than the background, as shown in Figure~\ref{fig:teaser}.
\begin{figure*}[t]
\centering \label{fig:main-fig}
     \includegraphics[width=1\textwidth]{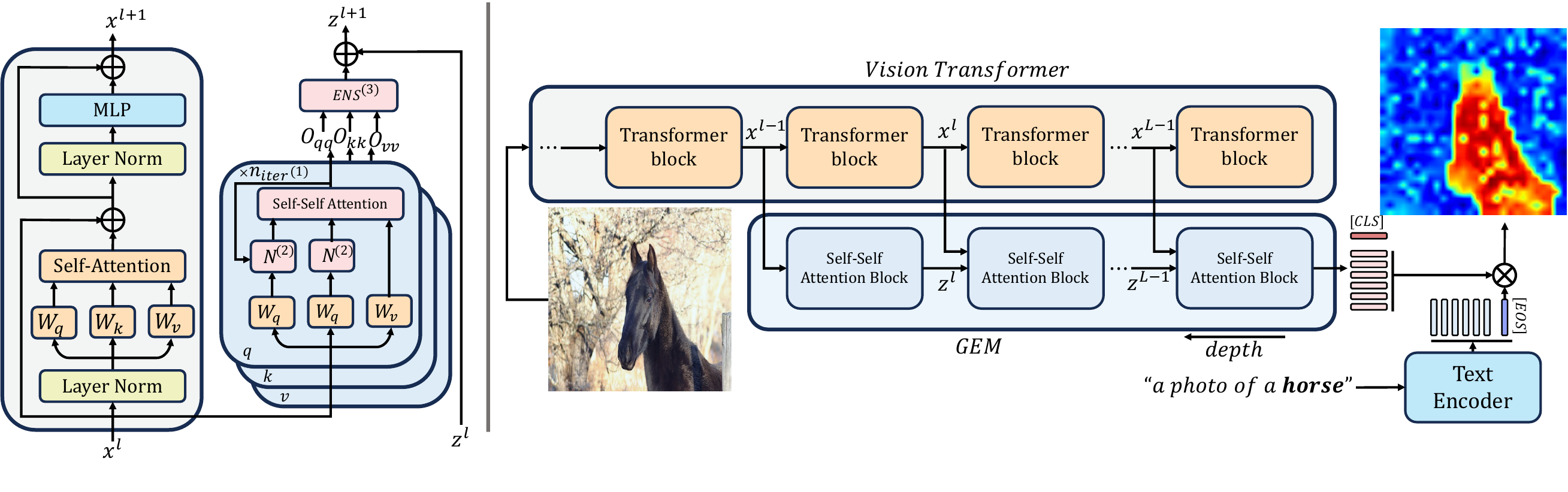}
     \vspace{-1cm}
     \caption{\textbf{Grounding Everything Module architecture:} (left) Overview of the proposed generalized self-self attention block including (1)iteration and (2)$L^2$ normalization $N$. The output of the q-q, k-k, and v-v projection is (3) ensembled before applying the skip connection.  (right) The output of self-self attention blocks is aggregated in parallel to the vision transformer in an alternative pathway. The localization is obtained by the dot product between the patch token output of the GEM and the CLS embedding of the text encoder. }
     \label{fig:overview}
     \vspace{-5mm}
\end{figure*}

In order to leverage vision-language models to localize objects in an open-vocabulary setting, different streams of approaches have been proposed. 
The first line of work trains a model to detect or segment regions in an image and then uses the vision-language information to label those regions as e.g. done in OVSeg\citep{liang2023open} or OpenSeg\citep{ghiasi2022scaling}.
A second line of work starts from the pretrained vision-language backbone and fine-tunes the model to improve localization, e.g. PACL~\citep{mukhoti2023open} or SegCLIP \citep{luo2023segclip}.
In contrast to that, a third line of work recently emerged that focuses on leveraging the inherent localization capabilities of models trained on image-level objectives without the need for annotations or retraining, namely
MaskCLIP~\citep{zhou2022extract} and CLIPSurgery \citep{li2023clipsurgery}.
Those training-free models try to process patches resp. tokens of the original model in a way that keeps them aligned to the language space and to avoid the inversion of image patch representation and text prompt. MaskCLIP showed that removing the MLP of the last layer avoids the vision-language inversion (see Figure~\ref{fig:teaser}). CLIPSurgery extends the pretrained ViT backbone of the CLIP model by a so-called ``surgery pathway'' which accumulates the value-value attentions of the original backbone over several layers. While adding the surgery pathway shows a significant performance improvement, it is not clear how this mechanism impacts the overall processing to achieve that improvement.

In this paper, we analyze the properties that result in the characteristics observed e.g. for CLIPSurgery and enforce them within a new, generalized self-self attention architecture. 
First, we show that the value-value attention can be generalized to a self-self attention, as any key-key, query-query, or value-value representations show similar characteristics.
Practically, we show that any form of self-self attention increases similarity among groups of similar tokens, compared to the standard q-k attention. 
To control the group formation, we propose a set of regularizations: first, we $\text{L}^{2}$ normalize the projected vectors; second, we combine this with an adaptive temperature $\tau$ for the proposed self-self attention operation, showing that the combination of those two elements results in good performance across all setups without the need for hyperparameter tuning.
Third, we show that repeating the self-self attention several times further increases the group formation. Finally, we ensemble over all self-self attention types to allow for an integration of all cues. 
An overview of the resulting Grounding Everything Module (GEM) architecture is shown in Figure~\ref{fig:overview}.

We evaluate the proposed method on two challenging tasks, open-vocabulary zero-shot semantic segmentation and zero-shot point prediction. For the first task, we leverage PascalVOC \citep{everingham2010pascal}, PascalContext \citep{mottaghi2014role}, as well as ADE20K~\citep{zhou2019sADE} dataset. For the second task, we employ the large-scale OpenImages V7 \citep{benenson2022colouring} dataset with almost 6K annotated classes. In all cases, we show improved results over all current training-free methods \citep{li2023clipsurgery, zhou2022extract} and competitive results in comparison to other approaches that require some form of fine-tuning \citep{xu2022groupvit, xu2023learning, luo2023segclip}.
It further shows that training-free methods in general and the proposed approach in particular are superior to all other approaches on the zero-shot point prediction on the OpenImages V7 dataset, reporting state-of-the-art results on this challenging task.

We summarize our contributions as follows:
(1) Inspired by \citet{li2023clipsurgery}, we show that self-self attention can be used as a technique for training-free open-vocabulary referential expression localization and segmentation based on pretrained vision-language models. 
(2) We propose the Grounding Everything Module (GEM) as a combination of self-self attention together with a set of regularizations that allows to generalize over a range of VL models and datasets.
(3) We provide an in-depth evaluation of our model and training-free methods in general, showing that they are able to keep up or even outperform fine-tuned methods on large-scale open vocabulary localization tasks.

\section{Related works}
\label{sec:related_works}

The success of large-scale vision-language models like CLIP has sparked interest in leveraging their abilities for tasks like open-vocabulary object localization.

Given the lack of ad-hoc localization properties of VL models, one line of approaches focuses on localization first e.g., by training a region-proposal detector or a segmentation network~\citep{kirillov2023segment}. They then use the respective vision-language models as a form of post-process labeling by computing the correlation of the respective regions with the text prompt. 
A representative example is OpenSeg~\citep{ghiasi2022scaling} that fine-tunes a model using class-agnostic masks and image-text pair data based on ALIGN~\citep{jia2021scaling}. 
Similarly, OVSeg consists of one segmentation model trained to generate class-agnostic masks in an open-vocabulary fashion, and one CLIP model adapted to classify these masks. MaskCLIP$^{(3)}$~\citep{ding2022open} adopts a similar strategy by using a Class-Agnostic Mask Proposal Network followed by a visual encoder based on CLIP to both refine the mask prediction and classify it.
By relying on a localization model with a closed set vocabulary, i.e., not trained on a web-scale dataset with a large vocabulary, the classification performance is focused on the vocabulary of that model.
Recently, GroundingSAM was proposed as a combination of GroundingDINO~\citep{liu2023grounding}, a model that leverages various sources of region-level supervision, such as masks and bounding boxes available for different vision tasks to train a general-purpose localizer, and SAM~\citep{kirillov2023segment} to generate segmentation masks from the bounding boxes generated by GroundingDINO.
Combining the supervision from various tasks allows these models to be trained on millions of samples with fine-grained supervision, thus achieving good performance for a large set of tasks.

Alternatively, some works propose to adapt the vision-language model architecture and training process to favor the emergence of localization. SegCLIP~\citep{luo2023segclip} and GroupViT~\citep{xu2022groupvit} modify the ViT architecture by interleaving regular transformer blocks with grouping blocks that allow the grouping of semantically similar tokens into learnable group tokens used to compute the contrastive loss with the text. Similarly, ViL-Seg~\citep{liu2022open} and OVSegmentor~\citep{xu2023learning} respectively use online clustering and Slot Attention~\citep{locatello2020object} for grouping visual features into semantically coherent clusters and in addition exploit self-supervision for refinement. Alternatively, ReCo~\citep{liu2021bootstrapping} leverages a retrieval process to obtain finer supervision and PACL~\citep{mukhoti2023open} trains a decoder on top of CLIP with a grounding loss. 
While these methods use image-caption pairs as supervision, they require heavy filtering of the dataset, like extracting common nouns, which makes the dataset lose its free-form text characteristic. Thus, such approaches do not fully benefit from the vision-language models' large-scale characteristics.  

Some methods refrain from training and instead adapt the pretrained vision-language model to make them work on fine-grained localization tasks.
MaskCLIP \citep{zhou2022extract} proposes discarding the Multi-Layer Perceptron (MLP) of the last layer of the vision transformer and utilizing the final value projection to extract dense patch-level features. Building upon this concept, CLIPSurgery \citep{li2023clipsurgery} introduces a novel pathway called the "surgery pathway" that operates in parallel with the original vision transformer (ViT) backbone of the CLIP model. It employs value-value instead of query-key attention and aggregates the output of multiple layers via residual connection.
Following~\citep{zhou2022extract}, the value-value attention is directly used without a subsequent MLP. To localize an object based on an input label or referential expression, the distance is computed between the token output of the last layer and the respective text embedding. The proposed work builds upon this stream of work and not only extends the value-value attention to a normalized self-self attention but also provides an in-depth analysis of the inner workings of self-self attention.

\section{Grounding with Self-Self Attention}\label{subsec:GEM}
\begin{figure}[t]
\centering
      \includegraphics[width=0.48\textwidth]{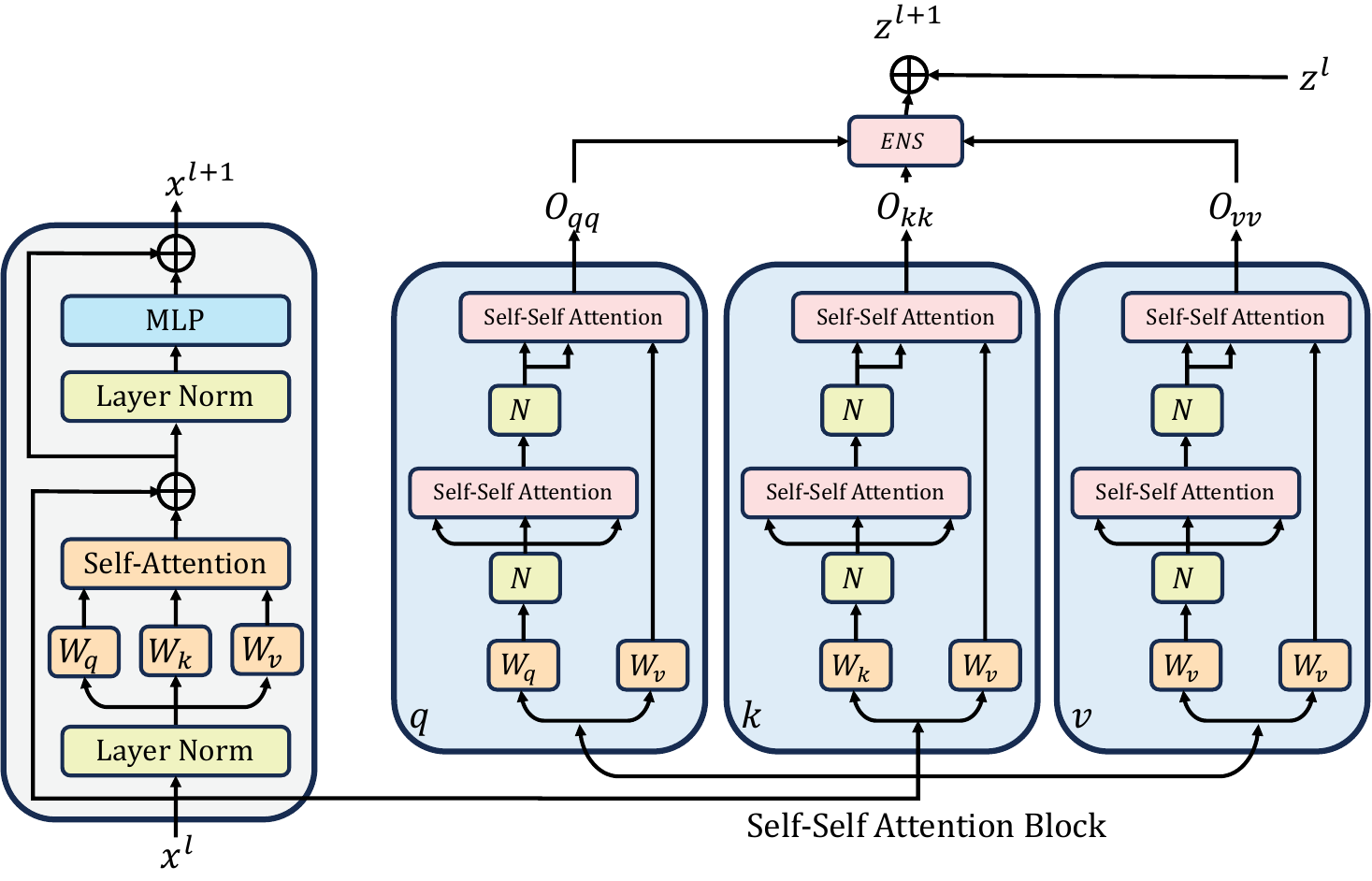}
     \caption{Detailed Illustration of GEM for a number of iterations for the iterative self-self attention equal to 1, where the block $N$ corresponds to $L^2$ normalization.}\label{fig:detailed-GEM}
\end{figure}
In the following, we introduce the Grounding Everything Module (GEM) by first generalizing the concept of value-value attention \citep{li2023clipsurgery} to a broader set of projections as self-self-attention and introduce an iterative extension that, together with a temperature regularizer, allows to control the formation of groups of visual features.
Second, we consider the connection of the proposed self-self attention (and also CLIPSurgery's value-value attention) to clustering, showing in simulations that it can act as a form of clustering. 
\subsection{GEM: Grounding Everything Module}\label{subsec:self-self-attn}
\paragraph{Self-Self Attention.}
We first review the concept of value-value attention, showing that, while it allows connecting features from the same semantic region, the same properties can be observed for key-key or query-query projections. CLIPSurgery defines value-value attention as:
\begin{equation}
Attn_{vv} = \mathrm{softmax}(V \cdot V^T), \quad O_{vv} = Attn_{vv} \cdot V
\end{equation}
with 
$V = x W_v \in R^{n\times d}$ with $x$ representing the patch tokens output by a ViT layer, $n$ represents the number and $d$ the dimension of tokens, respectively, and $W_v$ is the learned value weight matrix of the original ViT backbone, and $O_{vv}$ is the output of the value-value surgery block. 

As a first step, we replace the value projection by either the query or the key projection taken from the original pathway. 
We, therefore, introduce a generalized self-self attention $Attn_{ss}$ as extension of the value-value attention as:
\begin{equation}
Attn_{ss} = \mathrm{softmax}(x W_{proj} \cdot (x W_{proj})^T) 
\end{equation}
with $x \in R^{n \times d}$ again representing the patch tokens output by a ViT layer, and $W_{proj}$ being a projection matrix of the respective ViT layer $W_{proj}\in\{W_v, W_q, W_k\}$.
We evaluate the performance for each projection in Table \ref{tab:attn_types} on the Pascal VOC and Pascal Context datasets (for evaluation details see Section~\ref{sec:experiements_setup}). It shows that the query-query and key-key attention leads to the same or improved performance compared to value-value. Compared to regular self-attention (query-key attention) as used in the CLIP baseline, any self-self attention improves performances significantly. We discuss in Section \ref{subsec:ssa_clustering} that this can be attributed to self-self attention increasing the similarity of already similar patch tokens, thus leading to cluster formation.

\paragraph{Normalization and Adaptive Temperature.}
In the self-self attention setting, projected tokens with high norms might disproportionately influence other tokens, regardless of their similarity with other visual tokens.
We therefore propose an $L^2$-normalization for each projected token before computing self-self attention.  
We can further guide the cluster formation by introducing a temperature $\tau$ in the softmax formulation of the self-self attention $Attn_{ss}$ as:
\begin{equation}\label{eq:self-self-attn-output}
    \mathrm{softmax}(a,\tau) = \frac{e^{a_i \cdot a_j^T/\tau}}{\sum_l e^{a_i \cdot a_l^T/\tau}}    
\end{equation}
where, $\cdot$ is the dot product operation.
Assuming a zero-shot setting without access to labeled training or validation data, we aim to fix the temperature $\tau$ for the self-self attention so that it performs well without requiring hyperparameter tuning.
Therefore, we propose an adaptive temperature using the average norm of the visual tokens before projection times the temperature originally used to train ViT as 
\begin{equation}
\tau = \frac{N \cdot \sqrt{d}}{\sum_i ||x_i||_2},
\end{equation}
where $N$ is the number of visual tokens and $d$ the dimension of tokens, respectively.
This combination of normalization and adaptive temperature improves the group formation and thus the localization as shown in Table \ref{tab:attn_types}. Further details on temperature ablation are available in Section \ref{subsec:temp}. 

\begin{table}
\centering
\begin{tabular}{c c c c c}
\toprule 
Projection & Norm.+Temp.  & VOC & Context \\ 
\midrule 
CLIP    & - & 10.4 & 7.7 \\
\midrule 
v-v  & \color{lightgray}\ding{55}  & 41.9 & 30.5 \\
k-k & \color{lightgray}\ding{55}  &43.9& 31.0\\
q-q & \color{lightgray}\ding{55}  & 43.8& 30.8\\
\hline
qkv & \color{lightgray}\ding{55} & 43.1& 30.7\\
\midrule
v-v &\ding{51} & 44.4 & 31.9 \\
k-k &\ding{51} & 44.8& 32.0\\
q-q &\ding{51} &44.7 & 31.5\\
\hline
qkv &\ding{51} & 45.1 & 32.3 \\
\bottomrule
\end{tabular}
\caption{mIoU for v-v, k-k, and q-q attention and qkv ensemble on PascalVOC and PascalContext with and without L$^2$-Norm and adaptive temperature.
}
\label{tab:attn_types}
\end{table}

\paragraph{Iterative Self-Self Attention.}
We propose to iteratively apply the proposed normalized self-self attention to facilitate the gradual refinement of the cluster formation of semantically related visual tokens.
More formally, given input visual tokens denoted as $x \in R^{n \times d}$ and a projection matrix $W_{proj} \in R^{d \times d}$, the $k$-th iteration of our iterative self-self attention is described as:
\begin{equation}\label{eq:ss-attn-iter}
\begin{aligned}   
\left \{
 \begin{array}{lll}
    p^0 &= \frac{x W_{proj}}{||x W_{proj}||_2} \\
    p^{k\prime} &= \mathrm{softmax}(p^{k-1} \cdot (p^{k-1})^T, \tau) \cdot p^{k-1} \\
    p^k &= \frac{p^{k\prime}}{||p^k||_2}
 \end{array}
\right .
\end{aligned}
\end{equation}
where $p^0$ is the is the normalized projection input to the self-self attention operation, $p^k$ is the output of the $k$-st application of the self-self attention as described in Equation~\ref{eq:ss-attn-iter}, multiplied with the output of the $k-1$ iteration and divided by its norm. 
After $K$ iterations of self-self attention, the output (for the $W_{proj}$ projection), denoted $O_{ss}$, is obtained by applying the assignment to the values since they are trained to carry semantic information: 
\begin{equation}\label{eq:self-self-attn-output-iter}
    O_{ss} = \mathrm{softmax}(p^K \cdot (p^K)^T, \tau) \cdot V     
\end{equation}
Practically, we found that one additional iteration, so two successive self-self attentions, is sufficient for most cases. We therefore, fix the iterations to one throughout the paper and provide an ablation in Section \ref{subsec:temp}.

\paragraph{qkv-Ensemble.}
We finally ensemble the iterative self-self attention applied to the query, key, and value projections to integrate the information brought by the different projections. 
The output $O_{qkv}$ of the proposed qkv-ensemble attention is formally described as follows:
\begin{equation}\label{eq:ss-attn-ens-output}
     O_{qkv} = \frac{(O_{qq} + O_{kk} + O_{vv})}{3}
\end{equation}
where $O_{qq}, O_{kk}, O_{vv}$ are the outputs based on the respective projection matrices $W_q, W_k, W_v$.
Table \ref{tab:attn_types} shows the improvement achieved by ensembling over the three normalized projections (see Figure \ref{fig:detailed-GEM}).

\subsection{Self-Self Attention for Clustering}\label{subsec:ssa_clustering}
\begin{figure}
\centering
\includegraphics[width=\linewidth]{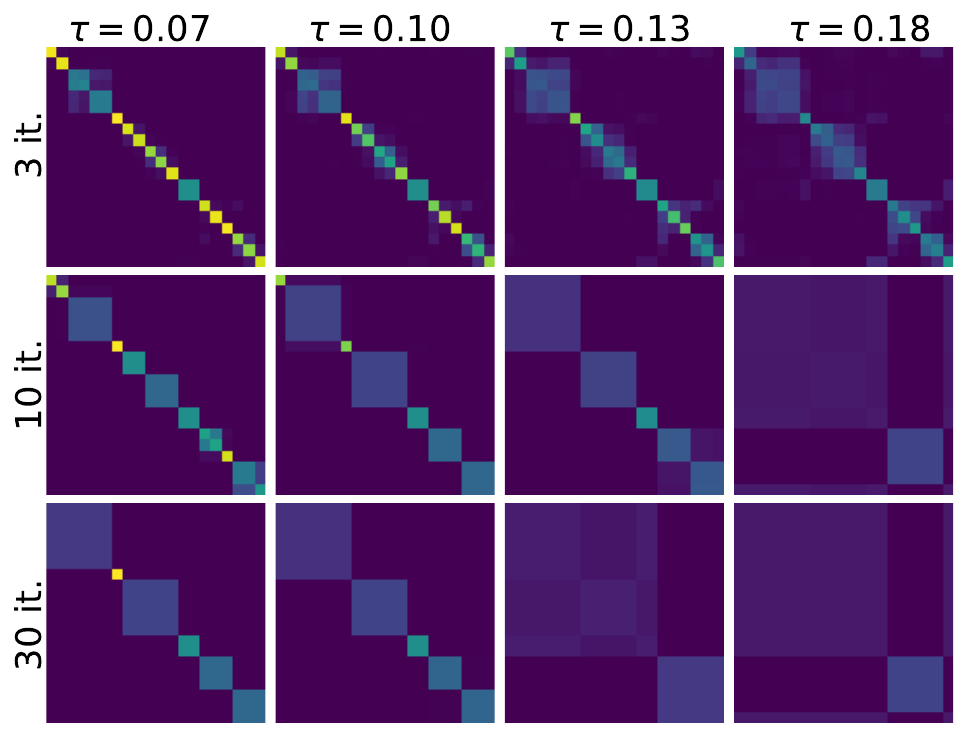}
\captionof{figure}{Evaluation of self-self attention for different numbers of iterations and temperature on a set of 20 random vectors (dim=5). It shows that as the number of iterations increases, self-self attention forms larger groups of clusters.
}
\label{fig:softmax-clustering}%
\vspace{-5mm}
\end{figure}

Practically, self-self attention calculates the similarity between each visual token and every other visual token. These similarities are then employed in the transformer as weights in a weighted sum operation used to update the tokens.
As a result, tokens are updated with a weighted sum of tokens, with more weight on more similar tokens, converging to a respective mean representation corresponding to a cluster center.
To validate this assumption, we conducted a simulation based on a set of 20 d-dimensional random Gaussian vectors representing the input token $x$ and a random linear projection as $W_{proj}$. We iteratively apply the proposed self-self-attention including normalization and with different temperature parameters on the 20 vectors.
As shown in Figure \ref{fig:softmax-clustering}, this process leads to a clustering of the 20 vectors using self-self attention. Moreover, it shows that higher temperature, as well as more iterations, lead to fewer, but larger clusters, while fewer iterations and a lower temperature enforce more and smaller clusters. 
In practical scenarios, complex datasets with many classes per image might benefit from a less clustered feature space, consequently requiring fewer iterations.

We can further connect this behavior to the Lipschitz constant of the used projections to the self-self attention's grouping effect.
More formally, in finite dimension, any linear operator is Lipschitz continuous and under the $l_2$ norm, its Lipschitz constant is given by the spectral norm of the weight matrix -- \textit{i.e.} the largest singular value of the weight matrix.
Let $W_{proj} \in R^{d \times d^\prime}$ denotes the weights matrix of the linear projection and $C$ its Lipschitz constant, we have:
\begin{equation}
\begin{aligned}    
 \forall x_1, x_2 \in R^{d}, &||x_2 W_{proj} - x_1W_{proj}||_2 \leq C || x_2 - x_1||_2 \\
 &C =  \max_{||x||_2 \neq 0} \frac{||xW_{proj}||_2}{||x||_2}
\end{aligned}
\end{equation}
For the self-self attention to reinforce the similarity of tokens already close to each other (\textit{i.e.} representing the same object), we need the self-self attention projection to pull these tokens closer to each other. In other words, the linear projection must be a contraction, \textit{i.e.} $C<1$. Conversely, a Lipschitz constant too small will result in having unrelated tokens to be mixed together. For the here analyzed models, we validated the Lipschitz constant across all projections as follows: $C_{value}=0.51 \pm 0.073$, $C_{key}=0.63 \pm 0.091$ and $C_{query}=0.66 \pm 0.104$.  
Moreover, the similarity between tokens (\textit{i.e.} grouping) in the self-self attention is further enforced by doing multiple (per head) parallel projections, all with a Lipschitz constant $<1$, as seen in value-value, query-query, or key-key projections.
Hence, tokens that are similar under all the projections will share information.
\section{Evaluation}\label{sec:experiements}

\begin{table*}

\centering
\resizebox{2\columnwidth}{!}{
\begin{tabular}{l|c|c|cc|c|c|ccc}
\toprule
 \multirow{2}{*}{Method} & \multirow{2}{*}{Encoder}& \multirow{2}{*}{Model} & \multicolumn{2}{c}{Dataset}  & Loc. & Loc. &\multicolumn{3}{c}{mIoU}\\
 &  &   & Pretraining & Annotation & anno. & FT & VOC & Context & ADE \\ \hline 
SPNet \citep{xian2019semantic}& ResNet101 & scratch &  COCO, VOC, Context & SM & \ding{51} & \ding{51} & 15.6$^\dag$ & 4.0$^\dag$ & -\\
ZS3Net \citep{bucher2019zero}& ResNet101 & scratch &  VOC, Context & SM & \ding{51} & \ding{51} & 17.7$^\dag$ & 7.7$^\dag$ & -\\
MaskCLIP$^{(3)}$ \citep{ding2022open} & ViT-B/16 & CLIP &  COCO & SM & \ding{51} & \ding{51} & - & 45.9 & 23.7\\ 
OpenSeg \citep{ghiasi2022scaling}& ENet-B7+FPN & ALIGN &  COCO, Loc. Narr & IT, UM  & \ding{51} & \ding{51} & 72.2 & 48.2& 24.8 \\
CLIP-ES \citep{lin2023clip} &  ResNet101 & CLIP &  COCO-Stuff-171 & IC & \ding{51} & \ding{51} & 75.0 & - & -\\ 
OVSeg \citep{liang2023open} & ViT-B/16 & CLIP &  COCO-Stuff-171 & UM & \ding{51}  & \ding{51} & 94.5 & 55.7 & 	29.6\\
\hline
\hline
ViL-Seg \citep{liu2022open}& ViT-B/16 & scratch & GCC & IT & \ding{55} & \ding{51} & 34.4$^\dag$ & 16.3$^\dag$ & -\\
GroupViT \citep{xu2022groupvit}& ViT-S/16 & scratch & GCC+YFCC & IT  & \ding{55} & \ding{51} & 52.3 & 22.4& 9.2\\
SegCLIP \citep{luo2023segclip}& ViT-B/16 & CLIP & CC, COCOcap & IT, ICap  & \ding{55} & \ding{51} & 52.6 & 24.7 &  8.7\\
OVSegmentor \citep{xu2023learning}& ViT-B/16 & DINO & GCC & IT  & \ding{55} & \ding{51} & 53.8 & 20.4& 5.6\\
\multirow{2}{*}{PACL \citep{mukhoti2023open}}  & \multirow{2}{*}{ViT-B/16} & \multirow{2}{*}{CLIP} &  WIT-400M & \multirow{2}{*}{IT } & \multirow{2}{*}{\ding{55}} & \multirow{2}{*}{\ding{51}} & \multirow{2}{*}{72.3} & \multirow{2}{*}{50.1} &\multirow{2}{*}{ 31.4}\\
                          &  &  &  +CC12M, YFCC &  & & & & \\ 
\hline
\hline
CLIP \citep{radford2021learning} & ViT-B/16 & CLIP &  WIT-400M & IT  & \ding{55} & \ding{55} & 10.4  & 7.7 & 1.7 \\
MaskCLIP$^{(2)}$ \citep{dong2023maskclip} & ViT-B/16 & scratch &   YFCC & IT  & \ding{55} & \ding{55} & - & 17.2 & 10.2 \\
MaskCLIP \citep{zhou2022extract} & ViT-B/16 & CLIP &  WIT-400M & IT  & \ding{55} & \ding{55} & - & 25.5 & - \\
MaskCLIP* \citep{zhou2022extract} & ViT-B/16 & CLIP &  WIT-400M & IT  & \ding{55} & \ding{55} & 28.6 & 23.8 &  10.2 \\
CLIP Surgery \citep{li2023clipsurgery} & ViT-B/16 & CLIP &  WIT-400M & IT  & \ding{55} & \ding{55} &  - & 29.3 & -\\ 
CLIP Surgery* \citep{li2023clipsurgery} & ViT-B/16 & CLIP &  WIT-400M & IT  & \ding{55} & \ding{55} &  \underline{41.2} & \underline{30.5}& \underline{12.9}\\ 
\hline
GEM (our)  & ViT-B/16 & CLIP &  WIT-400M  & IT  & \ding{55} & \ding{55} & \textbf{46.2}& \textbf{32.6}& \textbf{15.7} \\
GEM (our)  & ViT-B/16 & MetaCLIP &  metaclip-400M  & IT  & \ding{55} & \ding{55} & \textbf{46.8} &  \textbf{34.5} & \textbf{17.1}\\
\bottomrule
\end{tabular}
}
\caption{\textbf{Comparison on zero-shot semantic segmentation:} Models marked with $^\dag$ are evaluated under relaxed constraints, specifically on a subset of unseen classes. * signify our evaluation. We use the following short form, COCO: COCO2017,  GCC: Google Conceptual Captions 12M, YFCC: YFCC15M, CC: Conceptual Captions, COCOCap: COCO Captions. SM: segmentation mask, IT: image-text, ICap: image caption, UM: unlabeled mask, IC: image classes.}
\label{tab:sota}
\end{table*}

\subsection{Setup}\label{sec:experiements_setup}
\paragraph{Datasets.}
\textbf{PascalVOC}~\citep{everingham2010pascal} provides segmentation masks for 20 classes in natural images, focusing on common objects like cats, dogs, cars, and airplanes.
An image contains 1.5 classes on average.
Following previous works \citep{li2023clipsurgery}, \citep{zhou2022extract}, we evaluate on the validation set.
\\
\textbf{Pascal Context}~\citep{mottaghi2014role} extends PascalVOC to 59 classes, supplemented by a background class. Compared to PascalVOC, it provides dense annotations for the whole scene. We evaluate on the test set, comprising of $5,104$ images with an average of 4.8 classes per image.
\\
\textbf{ADE20K} \citep{zhou2019sADE} is a scene parsing dataset with 150 fine-grained classes. We use its validation set comprising of $2000$ images with an average of $9.9$ classes per image.
\\
\textbf{OpenImages-V7}~\citep{benenson2022colouring} provides annotations for a large set of images with a widely diverse spectrum of objects and real-world scenarios.
For the following evaluation, we leverage the point-wise annotations of the validation set, with $36,702$ images featuring $5,827$ distinct class labels. 
For each object, a set of positive and negative point annotations is provided. 
For this evaluation, for each image, we consider only classes present in the image. 
\vspace{-5mm}
\paragraph{Implementation.}
For all experiments, we use the original pretrained weights of the respective vision-langauge models as provided by their authors, namely CLIP~\citep{radford2021learning}, OpenCLIP \citep{cherti2023reproducible}, an open-source replication of CLIP, and BLIP~\citep{li2022blip} and MetaCLIP\citep{xu2023demystifying}.
We apply the GEM architecture with the proposed normalization and adaptive temperature and one iteration for all datasets and models. 
We compute a dense semantic segmentation prediction for each image as follows: For each patch we compute the cosine similarity between the patch tokens of the vision encoder and the text embedding of each dataset class name.
We use the following prompt as input for the text encoder: \textit{"a photo of a \{class name\}"}. 
Finally, we upsample the segmentation predictions to the input image size via bilinear interpolation.
If the input image is larger than the one used during the model training, we adapt the learned positional embeddings via bicubic interpolation.
Note that \emph{\textbf{we do not perform any retraining nor fine-tuning}} of the vision-language model, showing the possibility to localize queries with models trained on image-level only and without the need for any localization information during training or fine-tuning.

\paragraph{Evaluation.}
Zero-shot segmentation entails the ability of a model to segment objects in an image without prior training on the evaluated classes.
Following common practice \citep{xu2022groupvit, luo2023segclip, xu2023learning}, we evaluate \textbf{zero-shot semantic segmentation} by the mean Intersection over Union (mIoU) for PascalVOC, PascalContext and ADE20K.
Following \citep{xu2022groupvit}, we resize each input image to have a shorter side length of 448.
For PascalVOC we predict only the foreground classes and get the background by thresholding the softmax-normalized-similarity between the patch tokens and the text embedding of each class name (using a fixed threshold of 0.85).
For Pascal Context, we follow common practice and evaluate only on the 59 foreground classes.
ADE20K provides a dense annotation and therefore does not consider background. 
For \textbf{zero-shot point prediction}, we leverage the OpenImages-V7 dataset. For each positive class in the image, we scale the prediction between zero and one 
and use a fixed threshold of $0.5$ to obtain the predicted mask. We follow the authors' guidelines \citep{benenson2022colouring} and compute the IoU over the sets of positive and negative ground-truth points for all classes in the respective image, denoted p-mIoU. 

\subsection{Comparison to State-of-the-art}
\paragraph{Zero-Shot Semantic Segmentation.}
We first compare the proposed approach for the task of zero-shot semantic segmentation. We consider three groups of state-of-the-art methods in open-vocabulary segmentation: First, we consider methods trained resp. fine-tuned with some form of labeling information, e.g. hand-annotated segmentation masks, such as OpenSeg~\citep{ghiasi2022scaling}, CLIP-RIS~\citep{yu2023zero}, MaskCLIP$^{(3)}$~\citep{ding2022open}, and OVSeg~\citep{liang2023open}. Note that most of those methods are trained on similar domains and vocabulary as the test datasets. 
Second, we report the performance of models trained explicitly for segmentation on image-caption pair annotations, i.e., GroupViT~\citep{xu2022groupvit}, OVSegmentor~\citep{xu2023learning}, SegCLIP~\citep{luo2023segclip}, and ViL-Seg~\citep{liu2022open}. While those methods do not use location annotation, they anyway fine-tune existing backbones for the task of localization.
We also consider PACL~\citep{mukhoti2023open} in this group, which trains a decoder on top of CLIP using a loss designed for patch grouping.
Finally, we directly compare against methods that perform training-free zero-shot segmentation, namely MaskCLIP, MaskCLIP$^{(2)}$, and CLIPSurgery.
We report the mIoU in Table~\ref{tab:sota}. 
It shows that the proposed method consistently outperforms all training-free approaches. 
It further shows that training-free methods are able to outperform vision-language models fine-tuned specifically for localization on the more complex dataset PascalContext and ADE20K surpassing all other models except PACL. 

\vspace{-5mm}
\paragraph{Zero-Shot Point Prediction:} 
To evaluate the true open-vocabulary qualities of the proposed method, we compare our method on the OpenImageV7 dataset with a vocabulary of almost 6k label classes to the strongest available trained or fine-tuned semantic segmentation models from Table~\ref{tab:sota}, namely OVSeg, SegCLIP,  and GroupViT, as well as to all training-free methods.
Table \ref{tab:V7-results} reports the p-mIoU and the inference speed for all methods. %
First, it shows that training-free methods i.e. GEM, CLIPSurgery and MaskCLIP, provide a significantly better performance than trained or fine-tuned methods supporting the intuition that fine-tuning on a smaller, but cleaner dataset reduces the vocabulary leading to lower performance on datasets with large vocabulary like OpenImagesV7 (see Section~\ref{appendix:qualitative_analysis} for qualitative comparison). 
For completeness, we also report numbers for the recently released GroundingSAM architecture~\citep{kirillov2023segment, liu2023grounding} that uses labeled bounding boxes and class-agnostic masks during training. To directly compare, we use the output of GEM to label masks generated by prompting SAM with a grid of points. It shows that even in this case, the proposed training-free method is able to outperform the fine-tuned GroundingSAM architecture.

\begin{table}
\resizebox{1.\linewidth}{!}{
\begin{tabular}{lccc|c}
\toprule
Method & Loc. anno. & Loc. FT & p-mIoU & fps\\ 
\hline 
OVSeg* \citep{liang2023open} & \ding{51} &  \ding{51} &  22.5 & 1.41\\
\hline
SegCLIP* \citep{luo2023segclip} & \ding{51} & \ding{55} & 32.1 & 21.39\\
GroupViT* \citep{xu2022groupvit} & \ding{51}&  \ding{55} & 36.5 & 24.61\\
\hline
\hline
CLIP \citep{radford2021learning} & \ding{55} & \ding{55} & 27.6 & 42.10\\
MaskCLIP* \citep{zhou2022extract} & \ding{55} & \ding{55} & 42.0 & 42.43\\
CLIPSurgery* \citep{li2023clipsurgery} & \ding{55} &  \ding{55} & \underline{47.8} & 38.47\\
\hline
GEM-CLIP (our) & \ding{55} &  \ding{55} & \textbf{50.9} & 37.24\\ 
GEM-MetaCLIP (our) & \ding{55} & \ding{55} & \textbf{51.9} & 37.24\\ 
\hline
\hline
GroundingSAM* \citep{kirillov2023segment, liu2023grounding} & \ding{51} & \ding{51} & \underline{53.3} & 0.59\\  
\hline
GEM-SAM-CLIP (our) & \ding{51} & \ding{55} &  \textbf{53.4} & 0.45\\ 
GEM-SAM-MetaCLIP (our) & \ding{51} & \ding{55} &  \textbf{55.2} & 0.45\\ 
\bottomrule
\end{tabular}
}
\caption{\textbf{Comparison on zero-shot point prediction:} We choose the best performing available approaches for ADE20K from Table~\ref{tab:sota} and apply them on the OpenImagesV7 dataset. We further report inference speed as fps for each model on one Nvidia A6000. }\label{tab:V7-results}
\vspace{-5mm}
\end{table}

\subsection{Abalation}

\begin{figure}[t] 
     \centering
     \begin{subfigure}[b]{0.45\textwidth}
         \centering
         \includegraphics[width=1.\textwidth]{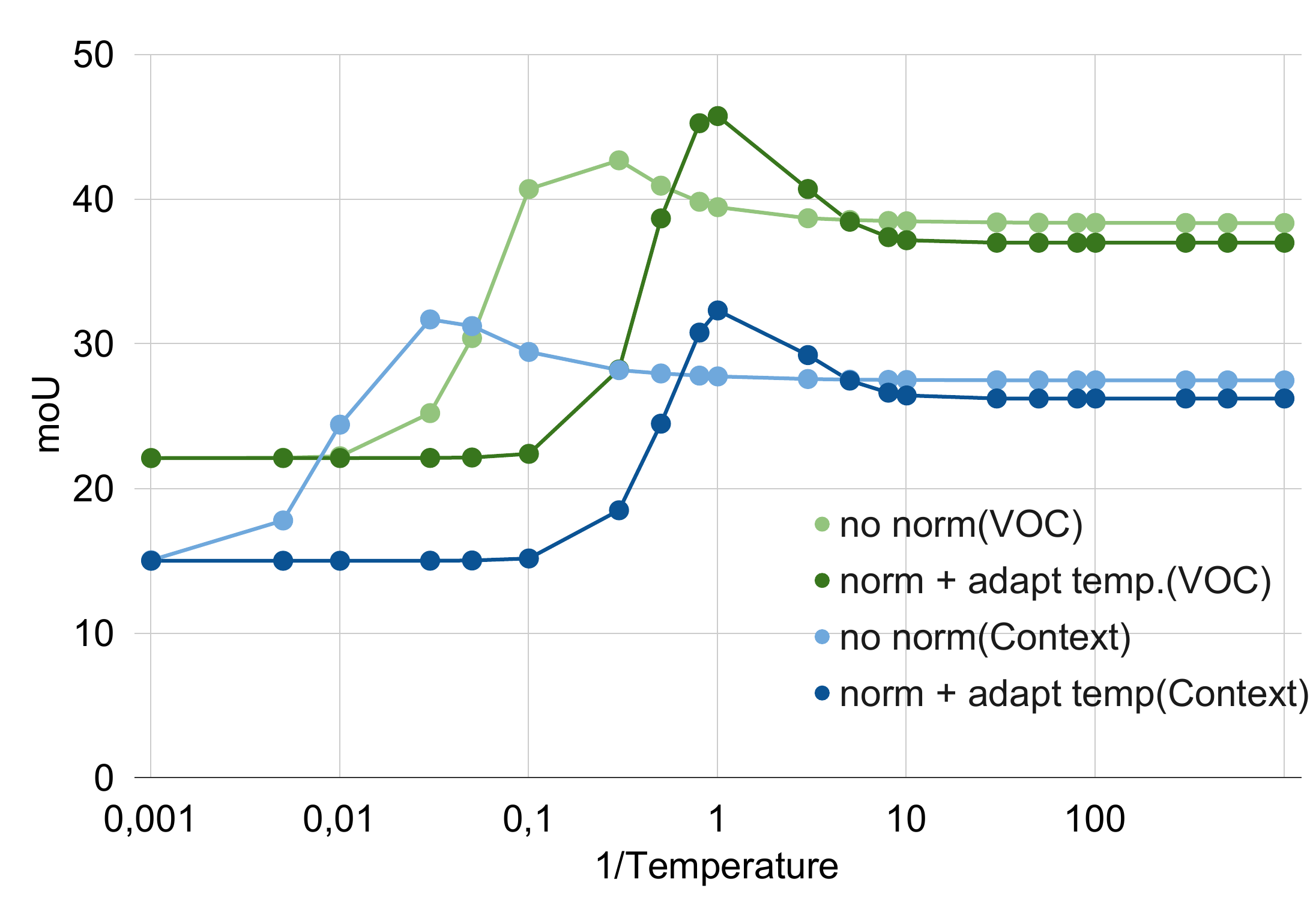}
    \end{subfigure}
    \vspace{-3mm}
    \caption{Evaluation of localization performance for CLIP ViT-B/16 (left) for the PascalVOC and PascalContext dataset with and without normalization and adaptive temperature. It shows that the proposed temperature provides best results in both settings.}\label{fig:temp-ablation}
\end{figure}

\paragraph{Temperature.} \label{subsec:temp}
To assess the performance of the proposed components, we first regard the impact of normalization and adaptive temperature. To this end, we compute the proposed adaptive temperature following in Section \ref{subsec:GEM}, i.e., $\tau = \frac{N \cdot \sqrt{d}}{\sum_i ||x_i||_2}$ and report the segmentation performance for multiples of this temperature for ViT-B/16 on two datasets, PascalVOC and PascalContext in Figure \ref{fig:temp-ablation}. 
We observe that the combination of normalization and temperature achieves the highest mIoU consistently across both datasets, but also that it achieved this performance consistently with the proposed temperature (multiplication factor equal to 1), indicating the effectiveness of our proposed heuristic as well as the robustness and generalizability, as it allows to adapt to the specific characteristics of the input vector. 
\vspace{-5mm}
\paragraph{Iterations.} Second we consider the impact of the number of iterations on the performance of the system. To this end, we evaluate PascalVOC and PascalContext for $K=\{0, 1, 2, 3\}$ iterations and also compare to the performance of the original CLIPSurgery pipeline in Table~\ref{tab:iterations}.  Overall, it shows that more iterations, namely two, slightly improve performance for VOC, a dataset with few classes per image, and that fewer iterations work slightly better for Context, a dataset with more classes per image. While the number of iterations can be used as a tunable hyperparameter, we fixed it throughout the paper at one to allow for a real zero-shot scenario. 

\begin{table}
\centering
\footnotesize
    \begin{tabular}{c | c | c c c c}
    \toprule 
    & CS & \multicolumn{4}{c}{GEM} \\ \hline
    iter  & 0 & 0& 1 & 2 & 3 \\ 
    \hline
    VOC     & 41.2 & 45.1 & \underline{45.5} & \textbf{46.2} & 45.6 \\ 
    Context & 30.5 & 31.5 & \textbf{32.6} & \underline{31.9} & 31.1\\ 
    \bottomrule
    \end{tabular}
    \caption{Influence of iterations for the self-self-attention in the GEM architecture. More iterations are better for fewer classes per image, less iterations work better for more classes. } \label{tab:iterations}
    \vspace{-5mm}
\end{table}

\begin{figure*}[t]
     \centering
     \begin{subfigure}[b]{0.3\textwidth}
         \centering
         \includegraphics[width=1\textwidth]{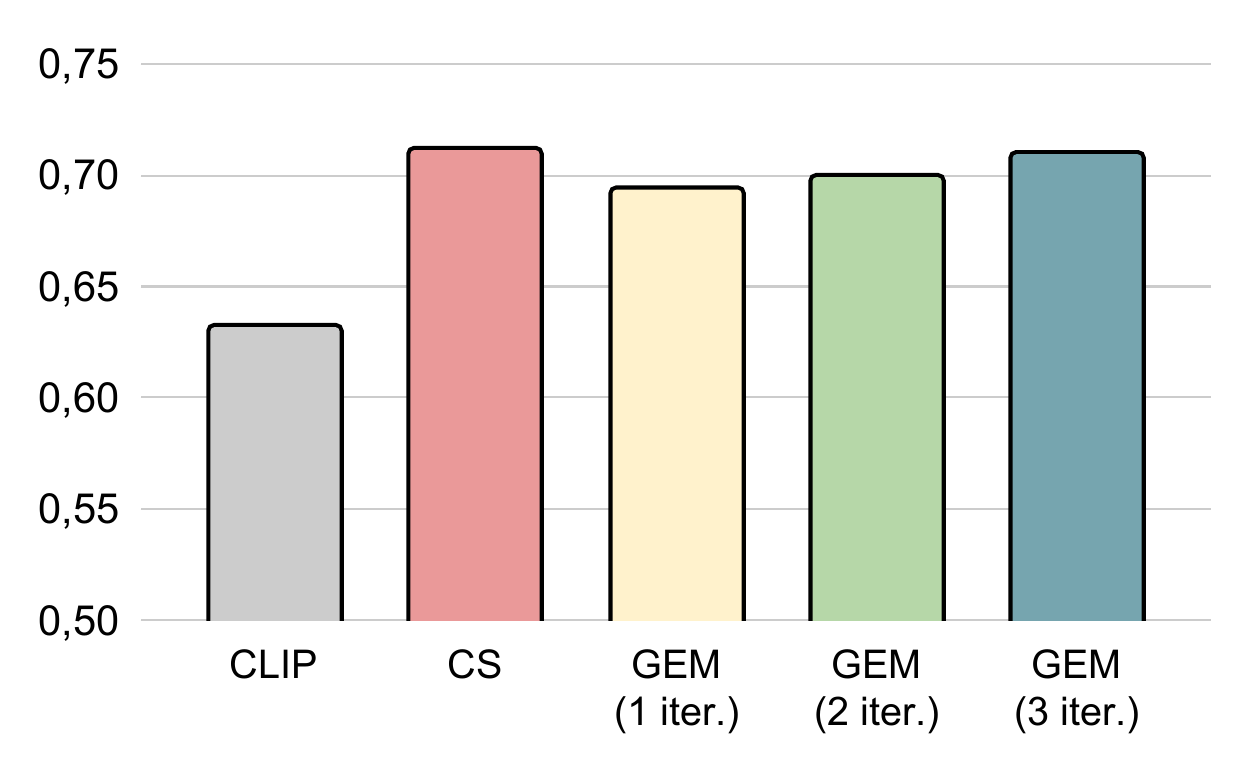}
         \caption{Patch-patch similarity}
         \label{fig:patch-patch-sim}
     \end{subfigure}
     \hfill
     \begin{subfigure}[b]{0.3\textwidth}
         \centering
         \includegraphics[width=1\textwidth]{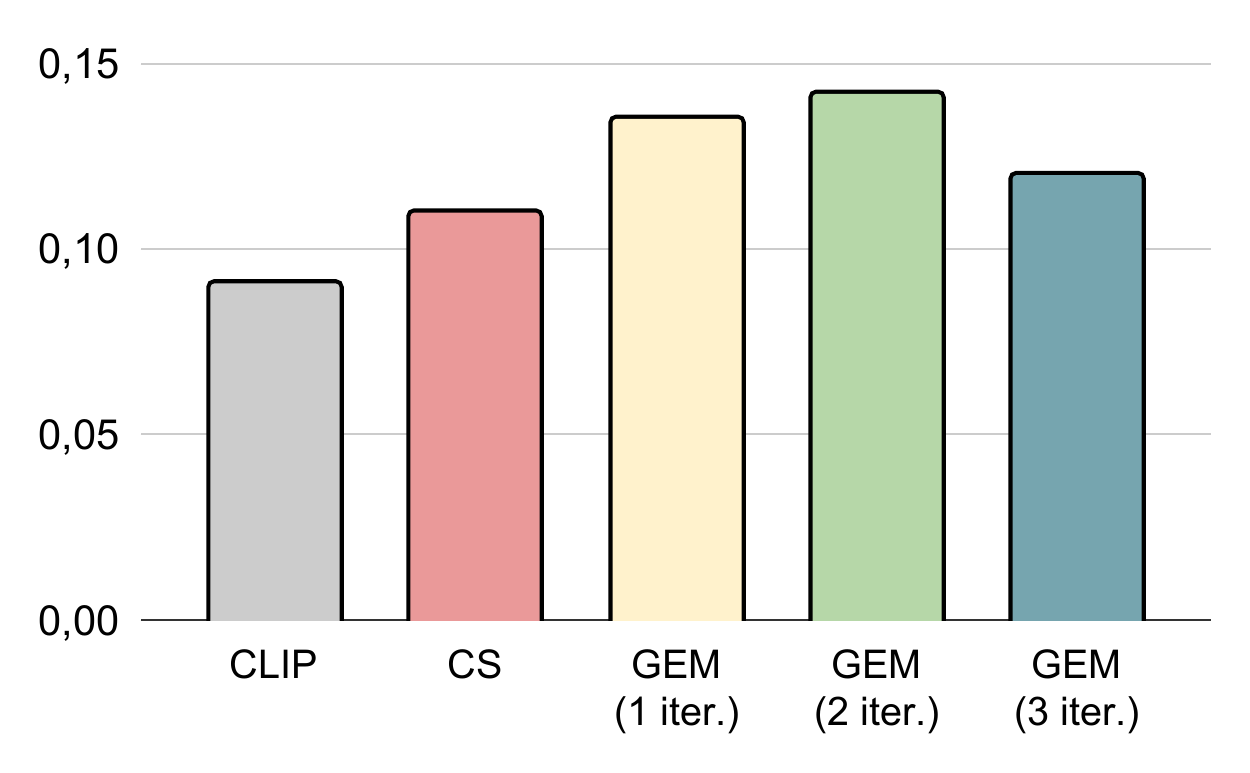}
         \caption{Object-Background Contrast}
         \label{fig:object-background-contrast}
     \end{subfigure}
     \hfill 
     \begin{subfigure}[b]{0.3\textwidth}
         \centering
         \includegraphics[width=1\textwidth]{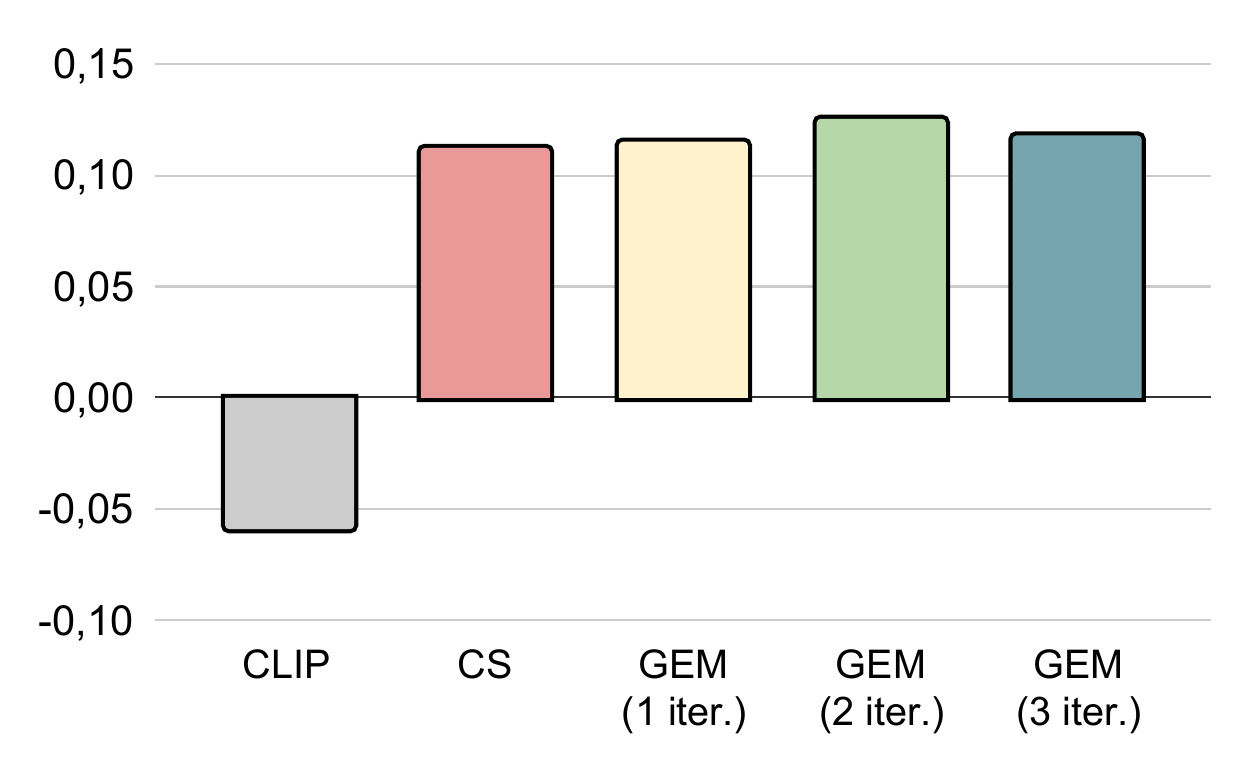}
         \caption{Text-Object Contrast}
         \label{fig:text-object-contrast}
     \end{subfigure}
     \vspace{-2mm}
     \caption{Metrics to analyze the localization properties of CLIP, CLIPSurgery, and our method GEM. Each metric is computed on the training set of the PascalVOC dataset.}
     \label{fig:metrics_visual_sim}
     \vspace{-2mm}
\end{figure*}

\subsection{Architecture and Model Size} \label{subsec:arch-size}

To explore the generalization abilities of the proposed method, we further extend our analysis beyond the ViT-B/16 model, including ViT-B/32 and ViT-L/14, as well as to other vision-language backbones, namely OpenCLIP~\citep{schuhmann2022laion},as an open-source replication of CLIP, thus to investigate the generality on an architecture closed to CLIP, BLIP~\citep{li2022blip}, as is trained with a multi-task objective, and MetaCLIP~\citep{xu2023demystifying} as the currently best-performing zero-shot classification model.
Table \ref{tab:model_sizes+architectures} shows the results for the different models and backbones. 
As expected, for a fixed ViT-B size, increasing the patch size from 16 to 32 reduces the performance slightly. We further observe that larger ViT-L encoders do not yield better localization performance. Specifically, GEM-ViT-B/16 consistently outperforms its larger counterparts GEM-ViT-L/14.
Finally, BLIP, as the only model trained with multi-objectives, tends to perform lower in localization than models trained solely with an image-text contrastive loss.
\begin{table}
\centering
\footnotesize
\begin{tabular}{c|c|ccc}
\toprule
Backbone & Model & VOC & Context & V7 \\ \hline 
\multirow{3}{*}{ViT-B/16}   &  CLIP &  \underline{46.2} &   \underline{32.6} &   \underline{50.9} \\
                            &OpenCLIP  & 43.1 &  31.7 &  49.9\\
                            &BLIP  & 42.8 &  23.5 & 45.2\\
                            &MetaCLIP  & \textbf{46.8} &   \textbf{34.5} & \textbf{51.9}\\
\hline 
\multirow{3}{*}{ViT-B/32}   &CLIP  & \textbf{40.5} & \underline{27.0} & \underline{46.6}\\
                            &OpenCLIP  & \underline{39.3} &  23.9 & 45.5\\
                            &MetaCLIP  & 38.2 &   \textbf{28.2} & \textbf{46.7} \\
\hline
\multirow{4}{*}{ViT-L/14}   &CLIP  & \underline{44.6} &  \textbf{28.6}& \textbf{46.3} \\
                            &OpenCLIP  & 40.0 &  \underline{27.5} & 42.4\\
                            &BLIP  & 32.1 &  21.4 & \underline{44.9}\\
                            &MetaCLIP  & \textbf{45.7} &  26.9 & 40.9\\
\bottomrule
\end{tabular}
\caption{Evaluation of the GEM architecture on various pretrained vision-language backbones showing better performance for smaller patch size (ViT-B/16 compared to ViT-B/32) and architecture (ViT-B compared to ViT-L).}\label{tab:model_sizes+architectures}
\vspace{-5mm}
\end{table}
\subsection{Analysis of Localization Properties}\label{sec:analysis}
In Figure \ref{fig:metrics_visual_sim}, we assess the factors contributing to the localization performance of the proposed method. We assume that for good localization in vision-language models, two essential properties must be fulfilled: visual distinctiveness as the meaningful grouping of visual feature representations, and vision-language alignment as the alignment of these groups with the textual descriptions encoded by the language model. 
To capture the visual distinctiveness, we consider two metrics: first, (a) patch-patch similarity, the similarity among patches within each layer, as well as, second, (b) object-background contrast, the contrast between foreground and background patch tokens. For this metric, we leverage the segmentation masks of the training set of the PascalVOC dataset \citep{everingham2010pascal}.
For vision-language alignment, (c), we measure the contrast between the similarity of the text embedding, the text-[EOS] token, and the foreground patch embeddings, and the similarity of the text-[EOS] token and the background patches.

We see an increase in patch-patch similarity (a) from CLIP to CLIPSurgery most likely due to the clustering induced by the self-self attention and the slight decrease from CLIPSurgery to GEM due to the added normalization and temperature. 
This is recovered by the higher object-background contrast (b) of GEM over CLIPSurgery and CLIP, pointing to the effective clustering of visual tokens and their ability to distinguish between distinct objects. 
Finally, the analysis of text-object similarity demonstrates improved alignment between visual tokens and text embeddings, enhancing vision-language integration. 
\subsection{Qualitative Analysis}\label{appendix:qualitative_analysis}
\begin{figure}[t]
\centering
      \includegraphics[width=0.49\textwidth]{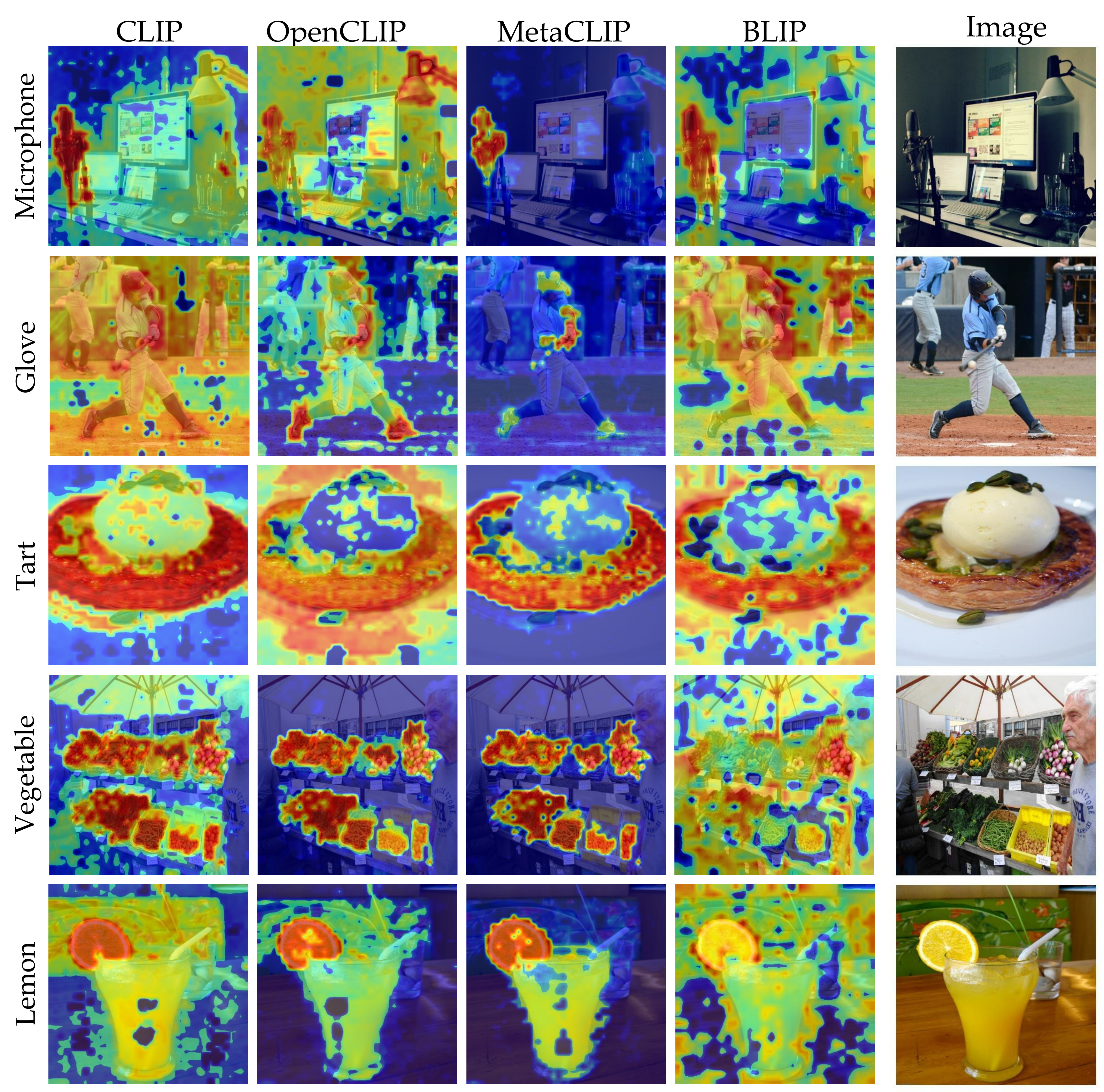}
     \caption{Qualitative comparison of GEM applied to different Vision-Language models.}\label{fig:GEMs_comparison}
\end{figure}
\begin{figure}
  \begin{center}
    \includegraphics[width=0.49\textwidth]{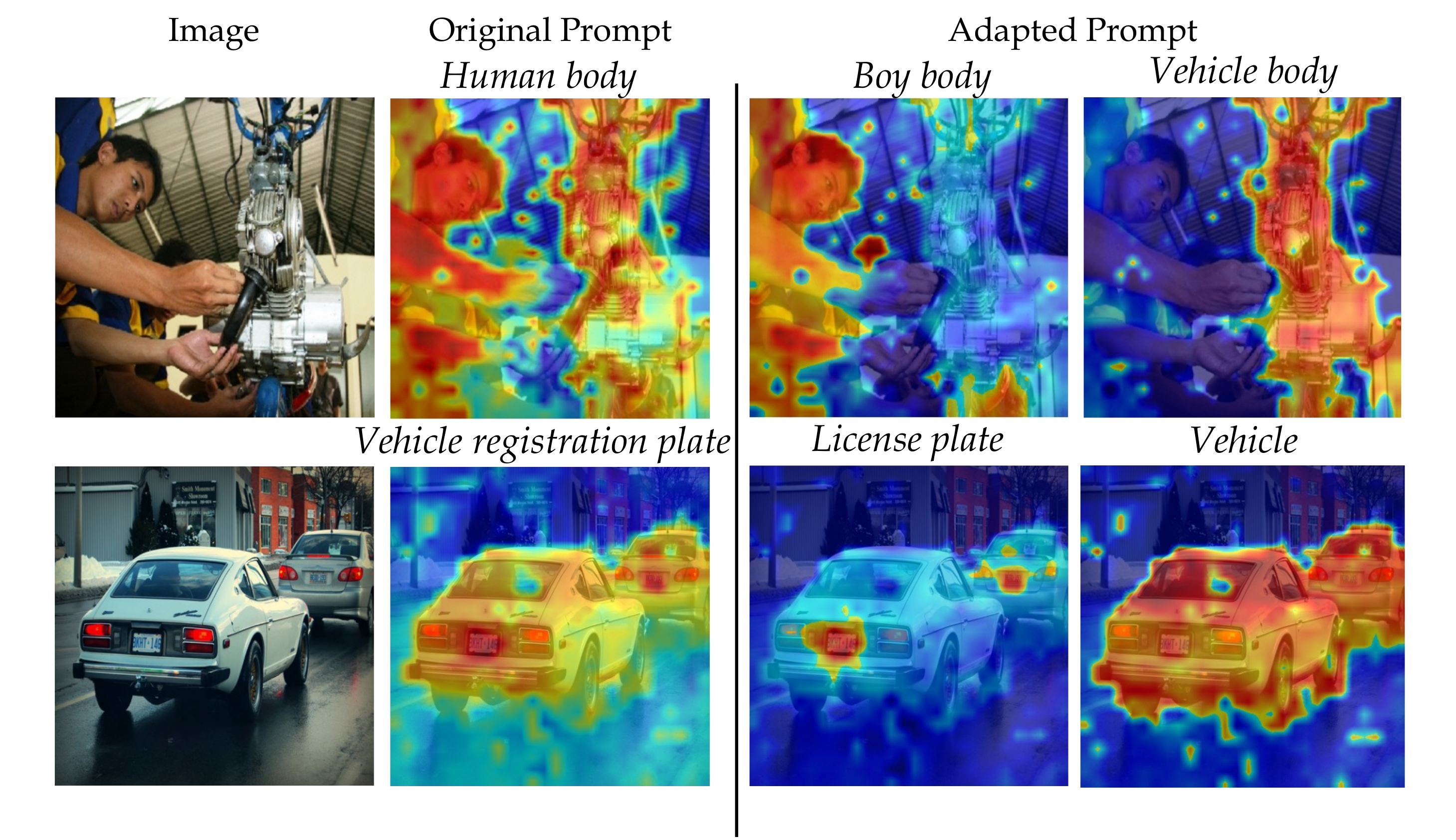}
  \end{center}
  \vspace{-7mm}
  \caption{Failure cases and adapted prompts from the OpenImagesV7 dataset.}\label{fig:failure-cases}
\end{figure}
\begin{figure*}[t]
\centering
      \includegraphics[width=0.82\textwidth]{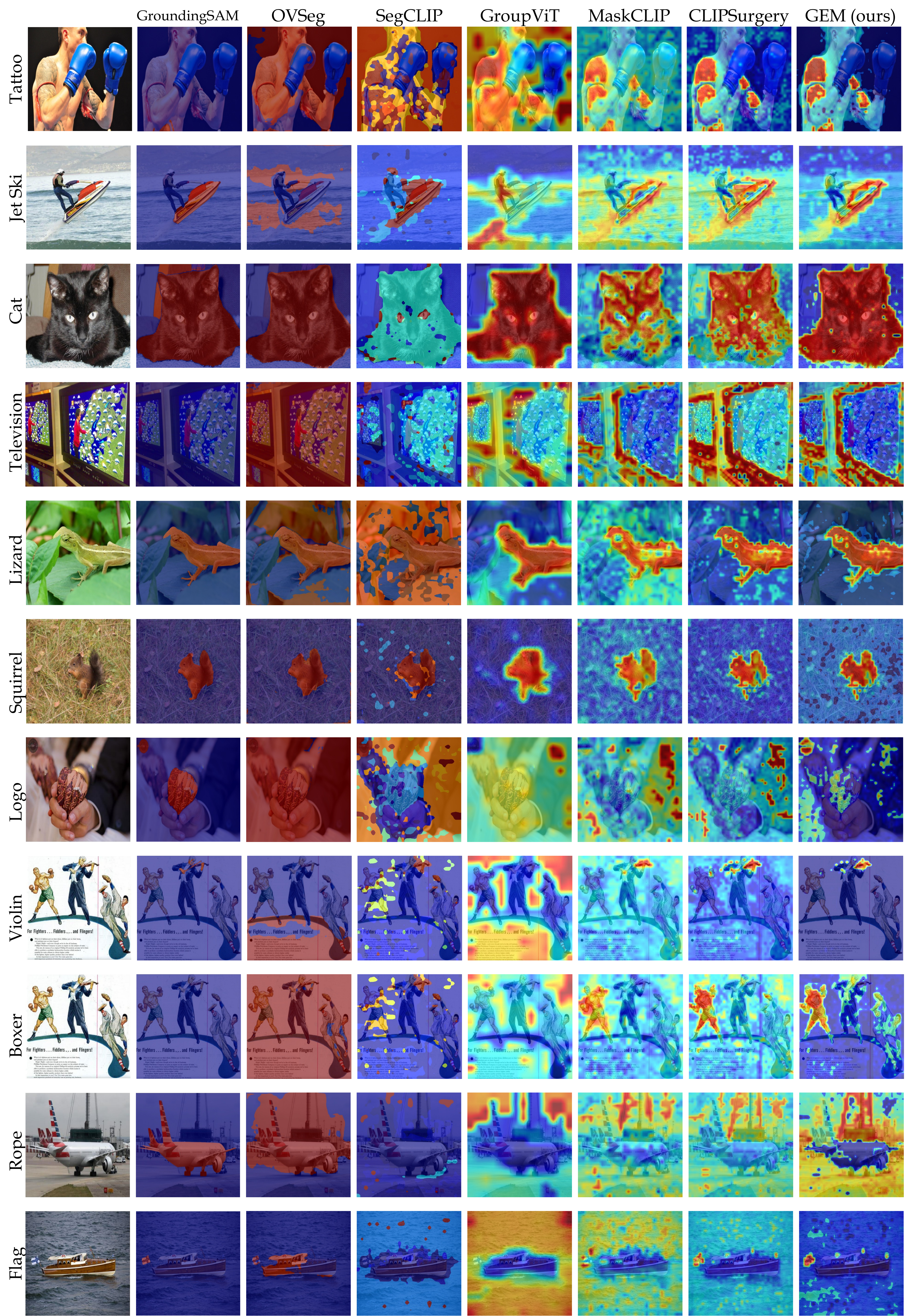}
     \caption{Qualitative comparison between different open-vocabulary segmentation methods, namely, GroundingSAM, OVSeg, SegCLIP, GroupViT, MaskCLIP, CLIPSurgery and GEM.}\label{fig:qualitative_comparison_big}
\end{figure*}

In the following, we discuss qualitative results for GEM:
\vspace{-1em}
\paragraph{Comparison of vision-language models}
Figure \ref{fig:GEMs_comparison} compares the localization performance of GEM applied to different vision-language models, namely, CLIP, OpenCLIP, MetaCLIP and BLIP. Overall, MetaCLIP produces sharper and more accurate localization compared to other models. It is also able to better identify objects, \textit{e.g.}, only GEM-MetaCLIP was able to localize "Glove" (Figure \ref{fig:GEMs_comparison} Row 2). Compared to that, GEM-BLIP, the only model trained with a multi-objective loss (contrastive, image-text matching and captioning) is still able to localize objects most of the time, but its segmentation mask is less precise. 
\vspace{-1em}
\paragraph{Analysis of Failure Cases}
Next, we review some failure cases in Figure~\ref{fig:failure-cases}. For the first image, when prompted with the text description \textit{''Human body''}, the model segments both the human and the car body. For the second image, prompted with \textit{''Vehicle registration plate''}, the model focuses again on both the car and registration plate. 
This effect can be mitigated by decoupling the emphasized word into  \textit{``Vehicle''} and \textit{``License plate''}, as shown in Figure \ref{fig:failure-cases}. We attribute this type of failure case to the text encoder, paving the way for future research.
\vspace{-1em}
\paragraph{Comparison to other methods}
Figure \ref{fig:qualitative_comparison_big} offers a qualitative comparison between different open-vocabulary segmentation methods. Included in the comparison are methods that use localization information (bounding box or mask) during training \textit{e.g.} GroundingSAM and OVSeg, that use a training strategy specifically tailored for segmentation \textit{e.g.} GroupViT and SegCLIP, and training-free methods \textit{e.g.} MaskCLIP, CLIPSurgery and our method GEM.

Figure \ref{fig:qualitative_comparison_big} shows that methods that were trained with localization information output high-quality masks (see "Cat", "Squirrel" and "Jet Ski") when the object is correctly identified. However, they are not able to detect entities in images that usually don't appear in detection and segmentation datasets. For example, neither GroundingSAM nor OVSeg are able the localize the "Boxer" or the "Violin" in the cartoon (Figure \ref{fig:qualitative_comparison_big} row 8 \& 9). This shows the limitation of using handcrafted segmentation annotation during training as they require too much effort to annotate and hence cover a much-restricted scope of entities. 

Methods that either fine-tune a pretrained Vision-Language like SegCLIP or train from scratch, are able to accurately segment common objects -- \textit{e.g.} "Cat" (Row 3), "Squirrel" (Row 6) and "Lizard" (Row 4) in Figure \ref{fig:qualitative_comparison_big} -- explaining the high performance they get on simply dataset like PascalVOC. However, these methods are unable to segment the rarest entities like the "Jet Ski" (Row 2), "Logo" (Row 7), or even the "Flag" (Row 11). We attribute this lack of diversity to their training strategy that involves the curation of the vocabulary of the used image-text pairs, therefore, reducing the size of the learned vocabulary.

Conversely, training-free methods like MaskCLIP, CLIPSurgery, and GEM benefit from the millions of image-text pairs that vision-language models are trained on, to be able to identify a diverse set of entities. While the segmentation masks of such models are not as sharp as the one outputted by GroundingSAM for example, they are able to localize objects like "Tattoo" (Row 1), "Television" (Row 4) and "Rope" (Row 10) that GroundingSAM is not able to localize. GEM outperforms its training-free counterparts in terms of segmentation sharpness (more defined contours and fewer holes) and is also able to localize objects missed by MaskCLIP and CLIPSurgery \textit{e.g.} "Logo" (Row 7).

\section{Conclusion}

In this work, we introduce the Grounding Everything Module, leveraging the latent localization capabilities of VL models trained on web-scale datasets. We propose a self-self attention pipeline for extracting localization information from vision-language models, complemented by a set of regularizations to ensure generalizability across diverse models and datasets, effectively enabling open-vocabulary localization without the need for additional training.


\section{Acknowledgment}
Walid Bousselham is supported by the German Federal Ministry of Education and Research (BMBF) project STCL - 01IS22067.
{
    \small
    \bibliographystyle{ieeenat_fullname}
    \bibliography{main}

\begin{thebibliography}{34}
\providecommand{\natexlab}[1]{#1}
\providecommand{\url}[1]{\texttt{#1}}
\expandafter\ifx\csname urlstyle\endcsname\relax
  \providecommand{\doi}[1]{doi: #1}\else
  \providecommand{\doi}{doi: \begingroup \urlstyle{rm}\Url}\fi

\bibitem[Benenson and Ferrari(2022)]{benenson2022colouring}
Rodrigo Benenson and Vittorio Ferrari.
\newblock From colouring-in to pointillism: revisiting semantic segmentation supervision.
\newblock \emph{arXiv preprint arXiv:2210.14142}, 2022.

\bibitem[Bucher et~al.(2019)Bucher, Vu, Cord, and P{\'e}rez]{bucher2019zero}
Maxime Bucher, Tuan-Hung Vu, Matthieu Cord, and Patrick P{\'e}rez.
\newblock Zero-shot semantic segmentation.
\newblock In \emph{NeurIPS}, 2019.

\bibitem[Cherti et~al.(2023)Cherti, Beaumont, Wightman, Wortsman, Ilharco, Gordon, Schuhmann, Schmidt, and Jitsev]{cherti2023reproducible}
Mehdi Cherti, Romain Beaumont, Ross Wightman, Mitchell Wortsman, Gabriel Ilharco, Cade Gordon, Christoph Schuhmann, Ludwig Schmidt, and Jenia Jitsev.
\newblock Reproducible scaling laws for contrastive language-image learning.
\newblock In \emph{CVPR}, 2023.

\bibitem[Ding et~al.(2022)Ding, Wang, and Tu]{ding2022open}
Zheng Ding, Jieke Wang, and Zhuowen Tu.
\newblock Open-vocabulary panoptic segmentation with maskclip.
\newblock In \emph{ICML}, 2022.

\bibitem[Dong et~al.(2023)Dong, Bao, Zheng, Zhang, Chen, Yang, Zeng, Zhang, Yuan, Chen, Wen, and Yu]{dong2023maskclip}
Xiaoyi Dong, Jianmin Bao, Yinglin Zheng, Ting Zhang, Dongdong Chen, Hao Yang, Ming Zeng, Weiming Zhang, Lu Yuan, Dong Chen, Fang Wen, and Nenghai Yu.
\newblock Maskclip: Masked self-distillation advances contrastive language-image pretraining.
\newblock In \emph{CVPR}, 2023.

\bibitem[Everingham et~al.(2010)Everingham, Van~Gool, Williams, Winn, and Zisserman]{everingham2010pascal}
Mark Everingham, Luc Van~Gool, Christopher~KI Williams, John Winn, and Andrew Zisserman.
\newblock The pascal visual object classes (voc) challenge.
\newblock In \emph{ICCV}, 2010.

\bibitem[Ghiasi et~al.(2022)Ghiasi, Gu, Cui, and Lin]{ghiasi2022scaling}
Golnaz Ghiasi, Xiuye Gu, Yin Cui, and Tsung-Yi Lin.
\newblock Scaling open-vocabulary image segmentation with image-level labels.
\newblock In \emph{ECCV}, 2022.

\bibitem[Jain et~al.(2021)Jain, Tancik, and Abbeel]{jain2021putting}
Ajay Jain, Matthew Tancik, and Pieter Abbeel.
\newblock Putting nerf on a diet: Semantically consistent few-shot view synthesis.
\newblock In \emph{ICCV}, 2021.

\bibitem[Jia et~al.(2021)Jia, Yang, Xia, Chen, Parekh, Pham, Le, Sung, Li, and Duerig]{jia2021scaling}
Chao Jia, Yinfei Yang, Ye Xia, Yi-Ting Chen, Zarana Parekh, Hieu Pham, Quoc Le, Yun-Hsuan Sung, Zhen Li, and Tom Duerig.
\newblock Scaling up visual and vision-language representation learning with noisy text supervision.
\newblock In \emph{ICML}, 2021.

\bibitem[Khan et~al.(2022)Khan, Kuehne, Gan, Lobo, and Shah]{khan2022weakly}
Aisha~Urooj Khan, Hilde Kuehne, Chuang Gan, Niels Da~Vitoria Lobo, and Mubarak Shah.
\newblock Weakly supervised grounding for vqa in vision-language transformers.
\newblock In \emph{ECCV}, 2022.

\bibitem[Kirillov et~al.(2023)Kirillov, Mintun, Ravi, Mao, Rolland, Gustafson, Xiao, Whitehead, Berg, Lo, et~al.]{kirillov2023segment}
Alexander Kirillov, Eric Mintun, Nikhila Ravi, Hanzi Mao, Chloe Rolland, Laura Gustafson, Tete Xiao, Spencer Whitehead, Alexander~C Berg, Wan-Yen Lo, et~al.
\newblock Segment anything.
\newblock In \emph{ICCV}, 2023.

\bibitem[Li et~al.(2022)Li, Li, Xiong, and Hoi]{li2022blip}
Junnan Li, Dongxu Li, Caiming Xiong, and Steven Hoi.
\newblock Blip: Bootstrapping language-image pre-training for unified vision-language understanding and generation.
\newblock In \emph{ICML}, 2022.

\bibitem[Li et~al.(2019)Li, Yatskar, Yin, Hsieh, and Chang]{li2019visualbert}
Liunian~Harold Li, Mark Yatskar, Da Yin, Cho-Jui Hsieh, and Kai-Wei Chang.
\newblock Visualbert: A simple and performant baseline for vision and language.
\newblock \emph{arXiv preprint arXiv:1908.03557}, 2019.

\bibitem[Li et~al.(2023)Li, Wang, Duan, and Li]{li2023clipsurgery}
Yi Li, Hualiang Wang, Yiqun Duan, and Xiaomeng Li.
\newblock Clip surgery for better explainability with enhancement in open-vocabulary tasks.
\newblock \emph{arXiv preprint arXiv:2304.05653}, 2023.

\bibitem[Liang et~al.(2023)Liang, Wu, Dai, Li, Zhao, Zhang, Zhang, Vajda, and Marculescu]{liang2023open}
Feng Liang, Bichen Wu, Xiaoliang Dai, Kunpeng Li, Yinan Zhao, Hang Zhang, Peizhao Zhang, Peter Vajda, and Diana Marculescu.
\newblock Open-vocabulary semantic segmentation with mask-adapted clip.
\newblock In \emph{CVPR}, 2023.

\bibitem[Lin et~al.(2023)Lin, Chen, Wang, Wu, Li, Lin, Liu, and He]{lin2023clip}
Yuqi Lin, Minghao Chen, Wenxiao Wang, Boxi Wu, Ke Li, Binbin Lin, Haifeng Liu, and Xiaofei He.
\newblock Clip is also an efficient segmenter: A text-driven approach for weakly supervised semantic segmentation.
\newblock In \emph{CVPR}, 2023.

\bibitem[Liu et~al.(2022)Liu, Wen, Han, Xu, Xu, and Liang]{liu2022open}
Quande Liu, Youpeng Wen, Jianhua Han, Chunjing Xu, Hang Xu, and Xiaodan Liang.
\newblock Open-world semantic segmentation via contrasting and clustering vision-language embedding.
\newblock In \emph{ECCV}, 2022.

\bibitem[Liu et~al.(2021)Liu, Zhi, Johns, and Davison]{liu2021bootstrapping}
Shikun Liu, Shuaifeng Zhi, Edward Johns, and Andrew Davison.
\newblock Bootstrapping semantic segmentation with regional contrast.
\newblock In \emph{ICLR}, 2021.

\bibitem[Liu et~al.(2023)Liu, Zeng, Ren, Li, Zhang, Yang, Li, Yang, Su, Zhu, et~al.]{liu2023grounding}
Shilong Liu, Zhaoyang Zeng, Tianhe Ren, Feng Li, Hao Zhang, Jie Yang, Chunyuan Li, Jianwei Yang, Hang Su, Jun Zhu, et~al.
\newblock Grounding dino: Marrying dino with grounded pre-training for open-set object detection.
\newblock \emph{arXiv preprint arXiv:2303.05499}, 2023.

\bibitem[Locatello et~al.(2020)Locatello, Weissenborn, Unterthiner, Mahendran, Heigold, Uszkoreit, Dosovitskiy, and Kipf]{locatello2020object}
Francesco Locatello, Dirk Weissenborn, Thomas Unterthiner, Aravindh Mahendran, Georg Heigold, Jakob Uszkoreit, Alexey Dosovitskiy, and Thomas Kipf.
\newblock Object-centric learning with slot attention.
\newblock In \emph{NeurIPS}, 2020.

\bibitem[Luo et~al.(2023)Luo, Bao, Wu, He, and Li]{luo2023segclip}
Huaishao Luo, Junwei Bao, Youzheng Wu, Xiaodong He, and Tianrui Li.
\newblock Segclip: Patch aggregation with learnable centers for open-vocabulary semantic segmentation.
\newblock In \emph{ICML}, 2023.

\bibitem[Mottaghi et~al.(2014)Mottaghi, Chen, Liu, Cho, Lee, Fidler, Urtasun, and Yuille]{mottaghi2014role}
Roozbeh Mottaghi, Xianjie Chen, Xiaobai Liu, Nam-Gyu Cho, Seong-Whan Lee, Sanja Fidler, Raquel Urtasun, and Alan Yuille.
\newblock The role of context for object detection and semantic segmentation in the wild.
\newblock In \emph{CVPR}, 2014.

\bibitem[Mukhoti et~al.(2023)Mukhoti, Lin, Poursaeed, Wang, Shah, Torr, and Lim]{mukhoti2023open}
Jishnu Mukhoti, Tsung-Yu Lin, Omid Poursaeed, Rui Wang, Ashish Shah, Philip~HS Torr, and Ser-Nam Lim.
\newblock Open vocabulary semantic segmentation with patch aligned contrastive learning.
\newblock In \emph{CVPR}, 2023.

\bibitem[Radford et~al.(2021)Radford, Kim, Hallacy, Ramesh, Goh, Agarwal, Sastry, Askell, Mishkin, Clark, et~al.]{radford2021learning}
Alec Radford, Jong~Wook Kim, Chris Hallacy, Aditya Ramesh, Gabriel Goh, Sandhini Agarwal, Girish Sastry, Amanda Askell, Pamela Mishkin, Jack Clark, et~al.
\newblock Learning transferable visual models from natural language supervision.
\newblock In \emph{ICML}, 2021.

\bibitem[Schuhmann et~al.(2022)Schuhmann, Beaumont, Vencu, Gordon, Wightman, Cherti, Coombes, Katta, Mullis, Wortsman, et~al.]{schuhmann2022laion}
Christoph Schuhmann, Romain Beaumont, Richard Vencu, Cade Gordon, Ross Wightman, Mehdi Cherti, Theo Coombes, Aarush Katta, Clayton Mullis, Mitchell Wortsman, et~al.
\newblock Laion-5b: An open large-scale dataset for training next generation image-text models.
\newblock In \emph{NeurIPS}, 2022.

\bibitem[Xian et~al.(2019)Xian, Choudhury, He, Schiele, and Akata]{xian2019semantic}
Yongqin Xian, Subhabrata Choudhury, Yang He, Bernt Schiele, and Zeynep Akata.
\newblock Semantic projection network for zero-and few-label semantic segmentation.
\newblock In \emph{CVPR}, 2019.

\bibitem[Xu et~al.(2023{\natexlab{a}})Xu, Xie, Tan, Huang, Howes, Sharma, Li, Ghosh, Zettlemoyer, and Feichtenhofer]{xu2023demystifying}
Hu Xu, Saining Xie, Xiaoqing~Ellen Tan, Po-Yao Huang, Russell Howes, Vasu Sharma, Shang-Wen Li, Gargi Ghosh, Luke Zettlemoyer, and Christoph Feichtenhofer.
\newblock Demystifying clip data.
\newblock \emph{arXiv preprint arXiv:2309.16671}, 2023{\natexlab{a}}.

\bibitem[Xu et~al.(2022)Xu, De~Mello, Liu, Byeon, Breuel, Kautz, and Wang]{xu2022groupvit}
Jiarui Xu, Shalini De~Mello, Sifei Liu, Wonmin Byeon, Thomas Breuel, Jan Kautz, and Xiaolong Wang.
\newblock Groupvit: Semantic segmentation emerges from text supervision.
\newblock In \emph{CVPR}, 2022.

\bibitem[Xu et~al.(2023{\natexlab{b}})Xu, Hou, Zhang, Feng, Wang, Qiao, and Xie]{xu2023learning}
Jilan Xu, Junlin Hou, Yuejie Zhang, Rui Feng, Yi Wang, Yu Qiao, and Weidi Xie.
\newblock Learning open-vocabulary semantic segmentation models from natural language supervision.
\newblock In \emph{CVPR}, 2023{\natexlab{b}}.

\bibitem[Yu et~al.(2022)Yu, Wang, Vasudevan, Yeung, Seyedhosseini, and Wu]{yu2022coca}
Jiahui Yu, Zirui Wang, Vijay Vasudevan, Legg Yeung, Mojtaba Seyedhosseini, and Yonghui Wu.
\newblock Coca: Contrastive captioners are image-text foundation models.
\newblock \emph{arXiv preprint arXiv:2205.01917}, 2022.

\bibitem[Yu et~al.(2023)Yu, Seo, and Son]{yu2023zero}
Seonghoon Yu, Paul~Hongsuck Seo, and Jeany Son.
\newblock Zero-shot referring image segmentation with global-local context features.
\newblock In \emph{CVPR}, 2023.

\bibitem[Yuan et~al.(2021)Yuan, Chen, Chen, Codella, Dai, Gao, Hu, Huang, Li, Li, et~al.]{yuan2021florence}
Lu Yuan, Dongdong Chen, Yi-Ling Chen, Noel Codella, Xiyang Dai, Jianfeng Gao, Houdong Hu, Xuedong Huang, Boxin Li, Chunyuan Li, et~al.
\newblock Florence: A new foundation model for computer vision.
\newblock \emph{arXiv preprint arXiv:2111.11432}, 2021.

\bibitem[Zhou et~al.(2019)Zhou, Zhao, Puig, Xiao, Fidler, Barriuso, and Torralba]{zhou2019sADE}
Bolei Zhou, Hang Zhao, Xavier Puig, Tete Xiao, Sanja Fidler, Adela Barriuso, and Antonio Torralba.
\newblock Semantic understanding of scenes through the ade20k dataset.
\newblock In \emph{International Journal of Computer Vision}, 2019.

\bibitem[Zhou et~al.(2022)Zhou, Loy, and Dai]{zhou2022extract}
Chong Zhou, Chen~Change Loy, and Bo Dai.
\newblock Extract free dense labels from clip.
\newblock In \emph{ECCV}, 2022.

\end{thebibliography}
}

\clearpage
\setcounter{page}{1}
\maketitlesupplementary

\noindent The supplementary material is organized as follows: We provide a link to a GoogleColab demo in Section~\ref{appendix:demo}. We then cover additional implementation details and present the rollout of one block in Section~\ref{appendix:implementation}. We further provide additional experimental ablation results in Section~\ref{appendix:ablation}. In Section~\ref{appendix:loc_prop_analysis}, we give more details on the analysis of localization properties and provide additional studies about those properties.
Finally, Section 10 provides additional details about the grouping factors.

\section{Colab Demo}
\label{appendix:demo}
We provide a GoogleColab demo at the following link: 

{\small \url{https://colab.research.google.com/drive/1f9aUbIpQIfEB8ZTUh3Krco8bIPqH3Pn3?usp=sharing}}


\begin{table*}[t]
\footnotesize
\begin{center}
\begin{tabular}{c c c ccccccccccc}
\toprule
\multirow{2}{*}{Backbone} & \multirow{2}{*}{MLP}& depth:  & 11 & 10 & 9 & 8 & 7 & 6 & 5 & 4 & 3 & 2 & 1\\ 
 & & layer:  & L2 & L3 & L4 & L5 & L6 & L7 & L8 & L9 & L10 & L11 & L12\\ 
\hline
ViT-B/16 & \ding{55} &  &  45.1 &45.4 &45.4 & 45.5& 45.5& 45.3 & \textbf{45.6} & \underline{45.5} & 45.2 & 43.8 & 4.8\\
ViT-B/16 & \color{lightgray} \ding{51} &   & 41.6 & 42.0 & 41.7 & 41.6 & 41.9 & \underline{42.3} & 42.2 & 42.1 & \textbf{42.4}  & 38.8 & 26.2 \\
\hline
ViT-B/32 & \ding{55} &  & 41.0	 &  41.0	 & 41.1 & 41.2 & 41.2 & 41.3 & \underline{41.4} & \textbf{41.5} & 40.3 & 26.1 & 5.1 \\
ViT-B/32 & \color{lightgray} \ding{51} &   & 39.7 & 39.6 & 39.7 & 40.0 & 40.1 & 40.1 & \underline{40.2} & \textbf{40.3} & 38.4 & 21.6 & 4.3 \\
\bottomrule
\end{tabular}
\end{center}
\vspace{-2em}
\caption{Evaluation of depth and impact of MLP on PascalVOC. We report mIoU performance depending on the depth resp. the starting layer of the self-self attention pipeline. It shows that starting at the middle layers provides best results, but also that higher layers can provide good results. In general, self-self attention without MLP outperforms self-self attention with MLP. }\label{tab:depth+mlp}
\vspace{-1.5em}
\end{table*}

\section{Additional Implementation Details}
\label{appendix:implementation}

GEM is built in parallel to the vision transformer by processing input features coming from the vision transformer through a series of ensembled iterative-temperature regularized self-self attention. We fix the number of iterations of self-self attention to one for all layers, i.e., we apply one step of self-self attention to the normalized projected features \textbf{and} one step of self-self attention to the values using the temperature heuristic as proposed section\ref{subsec:self-self-attn}. Figure \ref{fig:detailed-GEM} shows the rolled-out processing pipeline for self-self attention with one iteration and ensembled over queue-queue, key-key, and value-value attention. In the first iteration step self-self attention is computed on the respective query, key, or value projection following Equation~\ref{eq:ss-attn-iter} (main paper), followed by self-self attention of the respective projection applied to the value projection following Equation~\ref{eq:self-self-attn-output} (main paper). Finally, all three projections are ensembled following Equation~\ref{eq:ss-attn-ens-output} (main paper).

\section{Additional Ablation}\label{appendix:ablation}
To gain a deeper understanding of the factors influencing the performance of our method, we provide two additional ablations. Namely, we disentangle GEM's performance for the depth of the vision transformer at which we apply self-self-attention and evaluate the effect of adding the MLPs from the vision transformer encoder after the self-self attention in the alternative pathway.

\paragraph{Impact of path length:}
In Table \ref{tab:depth+mlp} we evaluate the segmentation performance of GEM applied to CLIP for two model sizes (ViT-B/16 and ViT-B/32) for different starting layers. We report the mIoU on PascalVOC. For both architectures, the performance remains significantly stable as long as GEM is applied before the last layers with best performance at a depth of three to five layers. We attribute the performance stability to the fact that the skip connections are essentially an exponential moving average applied at each layer. Therefore, the influence on the output features of the first layers decays exponentially. In general, we fix the depth $d$ of GEM to equal to $d=4$ for all reported experiments.

\paragraph{Impact of MLP:}
Originally, the studied vision-language models were trained using MLPs in their transformer blocks. While MaskCLIP~\citep{dong2023maskclip} and CLIPSurgery~\citep{li2023clipsurgery} already showed the negative impact of the MLP,  we further assess the influence of these MLPs on the downstream performance for the GEM architecture.
Table \ref{tab:depth+mlp} reports the mIoU on PascalVOC for ViT-B/16 and ViT-B/32 for different depths with and without the MLPs. We can see that adding MLPs have a slight negative effect on the downstream performance. While this is not a significant drop, it still shows that omitting MLPs will in general lead to better results.

\section{Further Details on Cluster Analysis} \label{appendix:analysis}

\begin{figure*}
    \centering
    \vspace{-1em}
    \includegraphics[width=\linewidth]{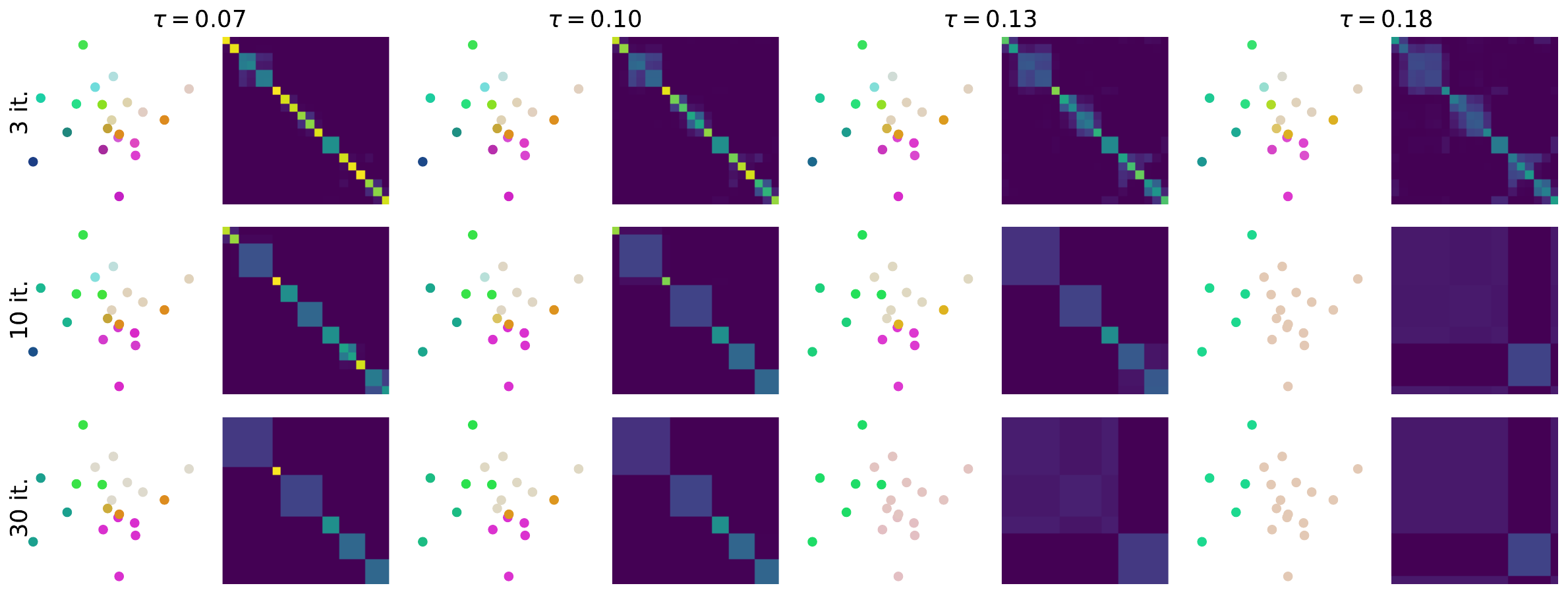}
    \vspace{-1.5em}
    \caption{
    Visualization of self-self attention on a set of 20 vectors: In the top 3 rows, a set of 20 vectors undergoing self-self attention for iterations $K=\{3, 10, 30\}$  and temperatures $\tau=\{0.07, 0.1, 0.13, 0.18\}$. 
    Displayed are the 20 data points (reduced to two dimensions via PCA) and their color represents a smooth cluster membership (the vector into which they are transformed is translated into a color value.) We further show the attention matrix for each configuration (the points were manually ordered for visual simplicity.) 
    It shows that as the number of iterations and/or the temperature increases, self-self attention produces larger fewer clusters.}
    \label{fig:ssa-clustering-full}
    \vspace{-1em}
\end{figure*}

Section \ref{subsec:ssa_clustering} discusses the idea that self-self attention acts as a form of clustering. In Figure \ref{fig:ssa-clustering-full} we extend the simulation presented in Section \ref{subsec:ssa_clustering} to more iterations and temperatures. We further add the point-cluster associations reduced to two dimensions via PCA to further visualize the cluster formation. In general, we can observe that increasing the number of iterations (from top to bottom) leads to fewer, larger clusters. The same holds for the temperature parameter where a higher temperature also leads to larger, fewer clusters.  

\section{Analysis of Localization Properties}\label{appendix:loc_prop_analysis}

In Section~\ref{sec:analysis}, we examine the factors contributing to the localization performance of the proposed method. In the following we provide details on the metrics used, a further discussion of the results as well as an analysis of those characteristics with respect to the depth of the GEM path. 
We assume that for localization in vision-language models, two essential properties must be fulfilled: \textit{visual distinctiveness}, which refers to the meaningful grouping of visual feature representations, and \textit{vision-language alignment}, which refers to the alignment of these groups with their respective textual descriptions encoded by the language model. In the case of CLIP, vision-language alignment translates to aligning patch tokens with the ViT [CLS] token, as the [CLS] token was trained to correlate with text embeddings through contrastive learning.

\subsection{Visual Distinctiveness} 
For visual distinctiveness, we consider two metrics:

\noindent \textbf{Patch-Patch Similarity.}
This captures the similarity among patches within each layer.  
We define an overall path-patch similarity as $S_{pp} = \frac{1}{n(n-1)} \sum\limits_{\substack{i, j \\ i \neq j}} x_i \cdot x_j^T$.

An increase in patch-patch similarity indicates a higher tendency for tokens to share similar characteristics.
However, high global path-patch similarity can also indicate that all patch tokens are near-identical, thus reducing localization effectiveness.


\noindent \textbf{Object-Background Contrast.}
We, therefore, further consider the object-background contrast. 
A critical characteristic of a model's localization proficiency is the ability to ensure similarity among patch tokens representing the same object while maintaining separation between those representing distinct objects. 
This characteristic permits the formation of semantically coherent clusters within the embedding space.
To this end, we adapt the Michelson contrast to measure the contrast in the similarity between foreground and background patch tokens. For this evaluation, we leverage the segmentation masks of the training set of the PascalVOC dataset \citep{everingham2010pascal}.
For a given segmentation mask $M$ of an object, we first compute the overall inside-to-inside similarity (noted $S^M_{in, in}$) and inside-to-outside ($S^M_{in, out}$): 

\begin{equation}
    \begin{aligned}
        S^{M}_{in, in} &= \frac{1}{m(m-1)} \sum\limits_{\substack{i, j \in M \\ i \neq j}} \cos(x_i, x_j)^+, \\
        S^{M}_{in, out} &= \frac{1}{m(n-m)} \sum\limits_{\substack{i\in M \\ k \notin \mathcal{M}}} \cos(x_i, x_k)^+ \\ 
    \end{aligned}
\end{equation}

Here, $m=|M|$ is the area covered by the mask, and the positive part function is employed to clamp negative similarities to zero, \textit{i.e.} $\cdot^+ = \max(0, \cdot)$. The object-background contrast ($C^M$) for an object mask $M$ is then defined as:
\begin{equation}
    C^M = \frac{S^M_{in, in} - S^M_{in, out}}{S^M_{in, in} + S^M_{in, out}}
\end{equation}

We average across all the masks in the dataset: $MC^M = \frac{1}{|\mathcal{M}|} \sum_{M \in \mathcal{M}} CS^M$,
with $|\mathcal{M}|$ being the total number of masks. Note that the ground truth masks are only used for analysis here.

\begin{figure}[t]
      \includegraphics[width=0.45\textwidth]{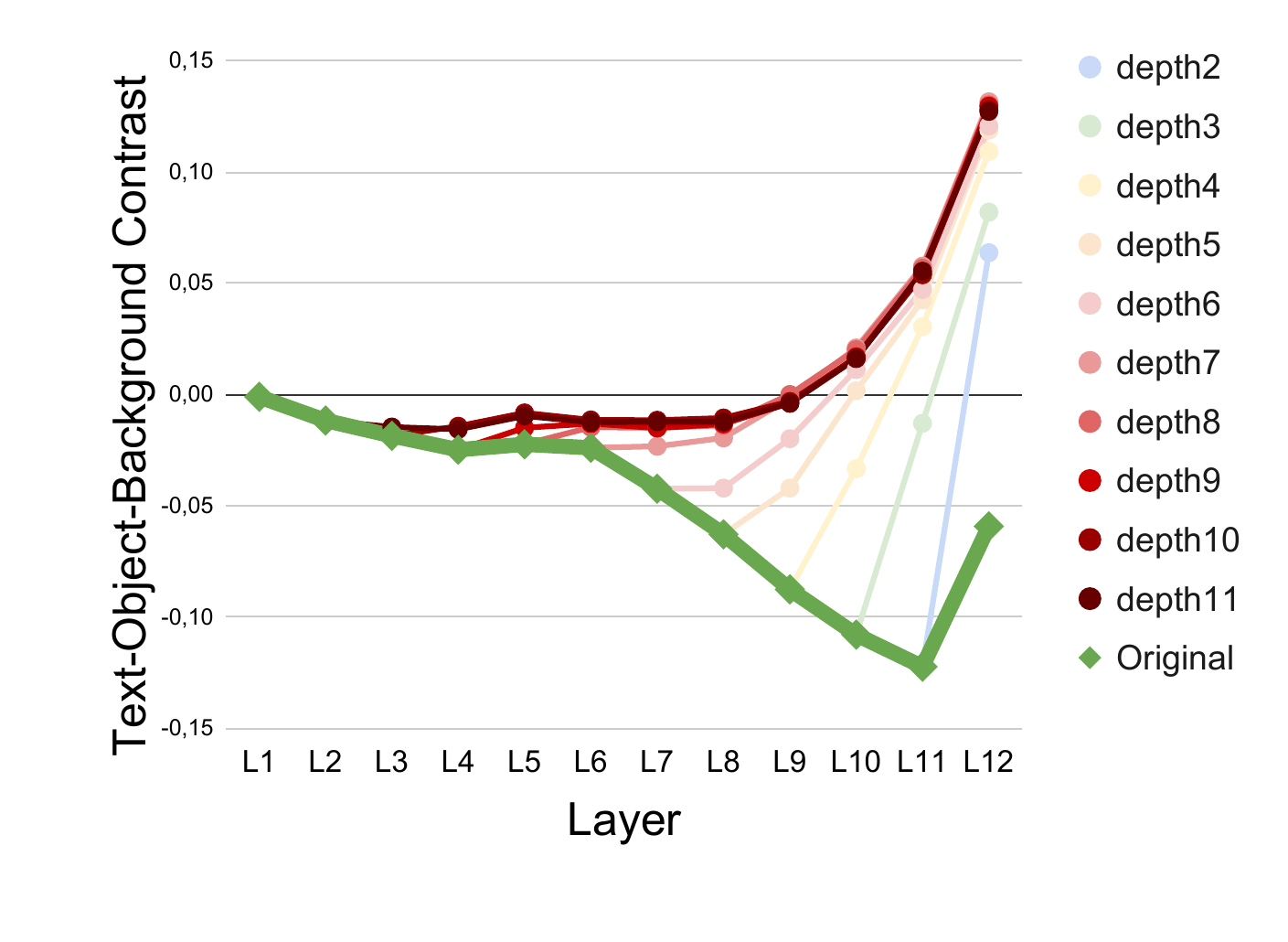}
      \vspace{-2em}
     \caption{Text-Object-Background contrast of CLIP (original) compared to GEM for different starting depth on PascalVOC for CLIP-ViT-B/16. }\label{fig:depth_txt_obj_bg_contrast_plots}
     \vspace{-2em}
\end{figure}

\subsection{Vision-Language Alignment} 

Second, we consider the problem of vision-language alignment.
Here, we aim to measure the contrast between the similarity of the text embedding representation of the class and the foreground patch embeddings, compared to the similarity of the text embedding and the background patches.

\noindent \textbf{Text-Object-Background contrast.}
%
Let $p \in R^{n \times d}$ be the patch token outputted by the vision transformer, where $n$ is the number of patches. For a segmentation mask $M$, the associated class name is denoted as $c(M)$, and we denote $t_{c(M)} \in R^{1 \times d}$ the text embedding of that class.
We compute the overall text-object similarity (noted $TS^M_{txt, obj}$) and text-background similarity ($S^M_{txt, bg}$): 

\begin{equation}
    \begin{aligned}
        TS^{M}_{txt, obj} &= \frac{1}{m} \sum\limits_{i\in M} \cos(t_{c(M)}, p_i)^+, \\
        TS^{M}_{txt, bg} &= \frac{1}{n-m} \sum\limits_{k \notin \mathcal{M}} \cos(t_{c(M)}, p_k)^+ \\
    \end{aligned}
\end{equation}

The text-object-background contrast for mask $M$ is then defined as: $TC^M = \frac{TS^M_{txt, obj} - TS^M_{txt, bg}}{TS^M_{txt, obj} + TS^M_{txt, bg}}$
This metric is subsequently averaged across all masks in the dataset to derive the global text-object-background contrast $MTC = \frac{1}{|\mathcal{M}|} \sum_{M \in \mathcal{M}} TC^M$.

A higher positive value for $MTC$ signifies that foreground patch embeddings are closer to their corresponding text embeddings than background patch embeddings. A negative value would indicate an inverse relationship.

\begin{figure}[t]
      \includegraphics[width=0.47\textwidth]{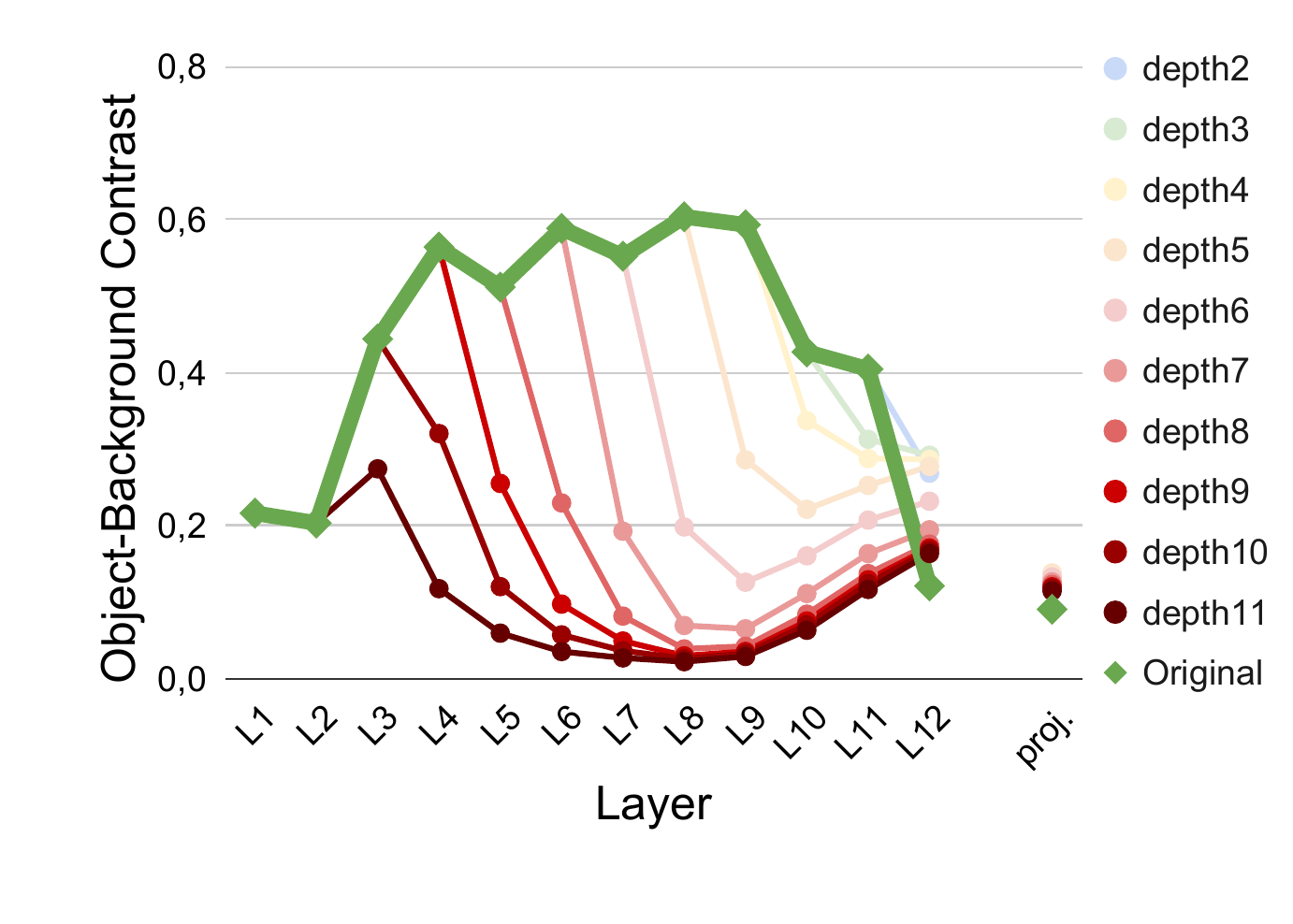}
      \vspace{-2.5em}
     \caption{Object-Background contrast of CLIP (original) compared to GEM for different starting depth on PascalVOC for CLIP-ViT-B/16. }\label{fig:depth_obj_bg_contrast_plots}
     \vspace{-2em}
\end{figure}

\subsection{Analysis}
Figure \ref{fig:metrics_visual_sim} in the main paper shows the results for the described metrics for CLIP, CLIPSurgery, and GEM for different numbers of iterations.
The observed increase in patch-patch similarity from CLIP to CLIPSurgery, in Figure \ref{fig:patch-patch-sim}, is due to the clustering induced by the self-self attention. 
We contribute the slight decrease for GEM to the added normalization and temperature. 
This is recovered by the higher object-background contrast of GEM over CLIPSurgery and CLIP, pointing to the effective grouping of visual tokens and their ability to distinguish between distinct objects.
Further, the analysis of text-object similarity demonstrates improved alignment between visual tokens and text embeddings, enhancing vision-language integration. 
Notably, CLIP, while exhibiting similar levels of visual distinctiveness in terms of patch-patch similarity and object-background contrast, significantly lags in terms of vision-language alignment, showing a negative text-object contrast, which means that background patches tend to align more closely with object-class text embeddings. This aligns with earlier findings in \citet{li2023clipsurgery} and \citet{mukhoti2023open}. 

We further analyze the impact of GEM with respect to the depth of the self-self attention as well as in comparison to the original model for a CLIP ViT/B-16 model on VOC. 
We show the object-background contrast (Figure~\ref{fig:depth_obj_bg_contrast_plots}) as well as the text-object-background contrast (Figure~\ref{fig:depth_txt_obj_bg_contrast_plots}) after each layer as well as for different depths. 
While the object-background contrast first drops by applying self-self attention, it also shows that it usually recovers after 3-4 layers, while the original CLIP architecture keeps a higher contrast, but significantly drops in the last three layers. 
Comparing this with the behavior of the text-object-background contrast (Figure~\ref{fig:depth_txt_obj_bg_contrast_plots}), we can see that the patch-language alignment of the original CLIP backbone drops significantly after layer six and only recovers in the last layer while the alignment of the self-self attention module consistently increases. Note that the original CLIP backbone always shows a negative text-object contrast, which means that background patches are more closely aligned to the object-class text embedding than the objects themselves while GEM reaches a positive alignment in the last layers.

\end{document}